\documentclass[11pt,a4paper]{article}
\usepackage[T1]{fontenc}
\usepackage[utf8]{inputenc}
\usepackage{lmodern}
\usepackage[margin=25mm]{geometry}
\usepackage{microtype}
\usepackage{amsmath,amssymb}
\usepackage{graphicx}
\usepackage{booktabs,tabularx,longtable}
\usepackage[round,authoryear]{natbib}
\usepackage[font=small,labelfont=bf]{caption}
\usepackage{placeins}
\usepackage{xcolor}
\usepackage{hyperref}
\hypersetup{colorlinks=true,linkcolor=black,citecolor=black,filecolor=black,urlcolor=blue,
  pdftitle={Learning Trajectories Leave Decodable Traces in Neural Network Weight Space and Internal Representations},
  pdfauthor={Patrick Krauss; Achim Schilling; Andreas Maier; Thomas Kinfe; Claus Metzner},
  pdfsubject={Initialization, learning history, and the limits of trajectory reconstruction}}
\graphicspath{{figures/}}
\newcommand{\R}{\mathbb{R}}
\newcommand{\norm}[1]{\left\lVert#1\right\rVert}
\newcommand{\CKA}{\operatorname{CKA}}
\newcommand{\relu}{\operatorname{ReLU}}
\newcommand{\suppref}[1]{\ref{#1}}

\title{Learning Trajectories Leave Decodable Traces in Neural Network Weight Space and Internal Representations}

\author{\begin{minipage}{\dimexpr\textwidth-2\tabcolsep\relax}
\centering\normalsize
Patrick Krauss\textsuperscript{1,2,3,4,*},
Achim Schilling\textsuperscript{2,3,4},
Andreas Maier\textsuperscript{1},\\
Thomas Kinfe\textsuperscript{2,4},
Claus Metzner\textsuperscript{1}\\[0.8em]
\footnotesize
\textsuperscript{1}Cognitive Computational Neuroscience Group, Pattern Recognition Lab,
Friedrich-Alexander-University Erlangen-N\"urnberg (FAU), Erlangen, Germany\endgraf
\textsuperscript{2}Neural Interface Lab, University Hospital Mannheim, University Heidelberg, Mannheim, Germany\endgraf
\textsuperscript{3}Neuroscience Lab, University Hospital Erlangen, Erlangen, Germany\endgraf
\textsuperscript{4}BG Clinik Ludwigshafen, Ludwigshafen am Rhein, Germany\endgraf
\smallskip
\textsuperscript{*}Correspondence: Patrick Krauss,
\href{mailto:patrick.krauss@fau.de}{patrick.krauss@fau.de}
\end{minipage}}
\date{}

\begin{document}
\maketitle
\vspace{-1.8em}
\begin{abstract}
Independently initialized neural networks trained on the same task cluster within the same region of high-dimensional weight space. However, their final weight configurations remain distinct even after extended training and do not converge to a single point in weight space. What these persistent differences represent, and what information they encode, has remained unclear. Are they merely alternative solutions to the same task, or do they encode traces of the individual learning histories that produced them? Here, we investigate whether final weight configurations retain information about their initial conditions and learning trajectories, and to what extent the trajectories leading to these final states can be reconstructed.
We find that training changed the geometry of neural network ensembles in weight space: pairwise distances between networks at initialization did not consistently predict the corresponding pairwise distances after training. Yet this geometric reorganization did not erase their individual origins. Remarkably, all 30 trained networks could still be correctly matched to their respective initial weight configurations among ten candidates per dataset. This historical signature extended even beyond weight space: networks sharing the same initialization developed similar final internal representations across different batch schedules, making their common origin identifiable from the learned representations themselves.
The temporal order of learning also left a decodable signature in the final weights. A linear probe recovered the order of earlier training tasks with 100 percent accuracy on 80 held-out networks, even after extended training on a common final task. The final weights thus provide partial access to a network’s past, but not a complete reconstruction of its learning trajectory.
Recovering a learned function does not require returning to the same location in weight space. When we perturbed network weights during training, the networks regained task performance without consistently returning to their original learning trajectories or converging toward the same final weight configurations as unperturbed networks. This dissociation was observed across datasets, perturbation stages, and network layers, demonstrating that functional recovery and geometric convergence in weight space are distinct.
Together, these findings show that networks trained on the same task can occupy a shared region of weight space while retaining distinct traces of their individual learning histories. Their final weight configurations contain decodable information about both their initial conditions and previous learning experiences. Thus, the final state of a neural network reflects not only what it has learned, but also how it arrived there.
\end{abstract}


\section{Introduction}


Can a trained neural network reveal its own past? Its present weights support a learned computation, but they were also produced by a particular initialization and a particular sequence of updates. Training could erase those individual beginnings as different networks adapt to a shared task. Alternatively, the solutions could retain recognizable traces of how they developed. Distinguishing these possibilities changes what we can learn from a final model: its structure may describe both the problem it solves and the history that produced it. In wide classification networks, class information can be distributed redundantly across neurons or concentrated in a sparse code \citep{vanmeegen2025coding}. Here we ask whether networks trained on the same final classification task also encode the order of earlier tasks.

Our preceding study showed that independently initialized networks form dataset- and task-dependent clusters in weight space after hidden-neuron permutations and positive ReLU rescalings are taken into account \citep{krauss2026clustering}. Related structure emerged in their internal representations, and a fixed-input comparison showed that changing the target rule redirected the organization. The separation began early, while displacement from initialization was still small. Nevertheless, networks solving the same problem did not collapse onto a common weight matrix. These findings established reproducible organization at the level of tasks, while leaving the origin of the remaining individual differences unresolved.

This raises a question that goes beyond the emergence of task-dependent organization: what explains the persistent differences between networks that have learned the same task? These differences could be incidental variations among functionally equivalent solutions. Alternatively, they could retain traces of the initial conditions and learning experiences that shaped each network. Distinguishing between these possibilities would reveal whether the geometry of weight space reflects only what a network has learned, or also how it learned it. This perspective extends our work on class separability, neural successor representations, and linguistic constructions \citep{schilling2021separability,stoewer2022successor,ramezani2025constructions}. Those studies characterize internal organization beyond predictive performance; the present study adds the history through which that organization develops.

Several forms of historical persistence must be distinguished. The distances among networks might preserve their ordering from initialization to the end of training. Alternatively, that ensemble geometry might be reorganized while each network retains an individually recognizable origin. Historical influence may also survive without direct resemblance to an earlier state: two copies of the same initialization might learn related final representations even when neither resembles its initial representation particularly closely. Finally, an endpoint might reveal a past event without containing enough accessible information to reconstruct the intervening sequence of weights. These possibilities require different tests.

Comparisons in weight space also require a definition of equivalence. Hidden neurons can be relabeled, and incoming and outgoing weights can be reciprocally rescaled in ReLU networks without changing their function. Permutation-aware matching makes it possible to compare organization beyond these coordinate freedoms \citep{ainsworth2023rebasin, krauss2026clustering}, while representation-based measures such as centered kernel alignment provide a complementary view of the responses to common inputs \citep{kornblith2019cka, krauss2026clustering}. Importantly, the geometry of weight space and the geometry of internal representations are not interchangeable. Differences in weight configurations do not necessarily imply corresponding differences in neural activations, nor do differences in internal representations necessarily imply differences in the underlying weights. Moreover, changes in representational geometry need not translate into corresponding changes in observable behavior, as recently demonstrated by Immertreu et al. (2026) \citep{immertreu2026geometry}.

Historical persistence also raises a dynamical question. Does a task-dependent cluster attract disturbed training trajectories, or can good performance be recovered along a different path? Research on recurrence resonance and information import in recurrent networks demonstrates how noise, state-space exploration, and computation can be related \citep{krauss2019recurrence,metzner2022dynamics}. Studies of reservoir computing and organizational regularities further separate useful computation from any single dynamical regime \citep{metzner2025reservoir,metzner2025organization}. The corresponding distinction in a feedforward learning system concerns the evolution of its parameters: recovery of classification need not entail recovery of the same weights.

We address these questions through a connected sequence of analyses and interventions. We first test whether initial geometry and individual identity persist together. We then vary initialization and batch schedule independently to identify their contributions to final weights and representations. Paired task histories test whether a common final objective erases the order of earlier learning. Controlled perturbations assess return toward an unperturbed continuation. Finally, endpoint probes test the boundary between recognizing a historical event and reconstructing a trajectory. Together, these experiments investigate to what extent the final configuration of a neural network reflects not only the task it has learned, but also the individual learning history through which it arrived there.

\section{Methods}
\label{sec:methods}

\subsection{Networks, datasets, and experimental cohorts}

All analyses used a fully connected network with 784 inputs, 64 ReLU hidden units, and ten outputs,
\begin{equation}
 f(x)=\relu(xW_0)W_1,
 \qquad W_0\in\R^{784\times64},\quad W_1\in\R^{64\times10}.
\end{equation}
Biases were fixed at zero. Weights used the default PyTorch linear-layer initialization, uniform within $\pm1/\sqrt{n_{\rm in}}$ for each layer's input dimension. Training minimized cross-entropy with stochastic gradient descent (SGD), learning rate 0.1, momentum 0.9, and batch size 256. Images were flattened after scaling pixel intensities to $[0,1]$ and standardizing with mean/standard-deviation pairs 0.1307/0.3081 for MNIST, 0.2860/0.3530 for Fashion-MNIST, and 0.1918/0.3483 for Kuzushiji-MNIST (KMNIST). Each dataset provided 60,000 training and 10,000 test images \citep{lecun1998mnist,xiao2017fashion,clanuwat2018kuzushiji}.

We initially analyzed 30 neural networks using the experimental setup of our previous study \citet{krauss2026clustering}. For each of the three datasets (MNIST, Fashion-MNIST, and KMNIST), ten independently initialized networks were trained for ten epochs. Their weights were saved before training and after every half epoch, resulting in 21 checkpoints per network (E0--E20). These data allowed us to compare the geometry of initial and final weight configurations and to investigate whether individual networks could be identified from their trained weights. Additional experiments used independently trained network ensembles to examine the effects of initialization, training batch order, task history, and weight perturbations.

\begin{figure}[tbp]
\centering\includegraphics[width=\linewidth]{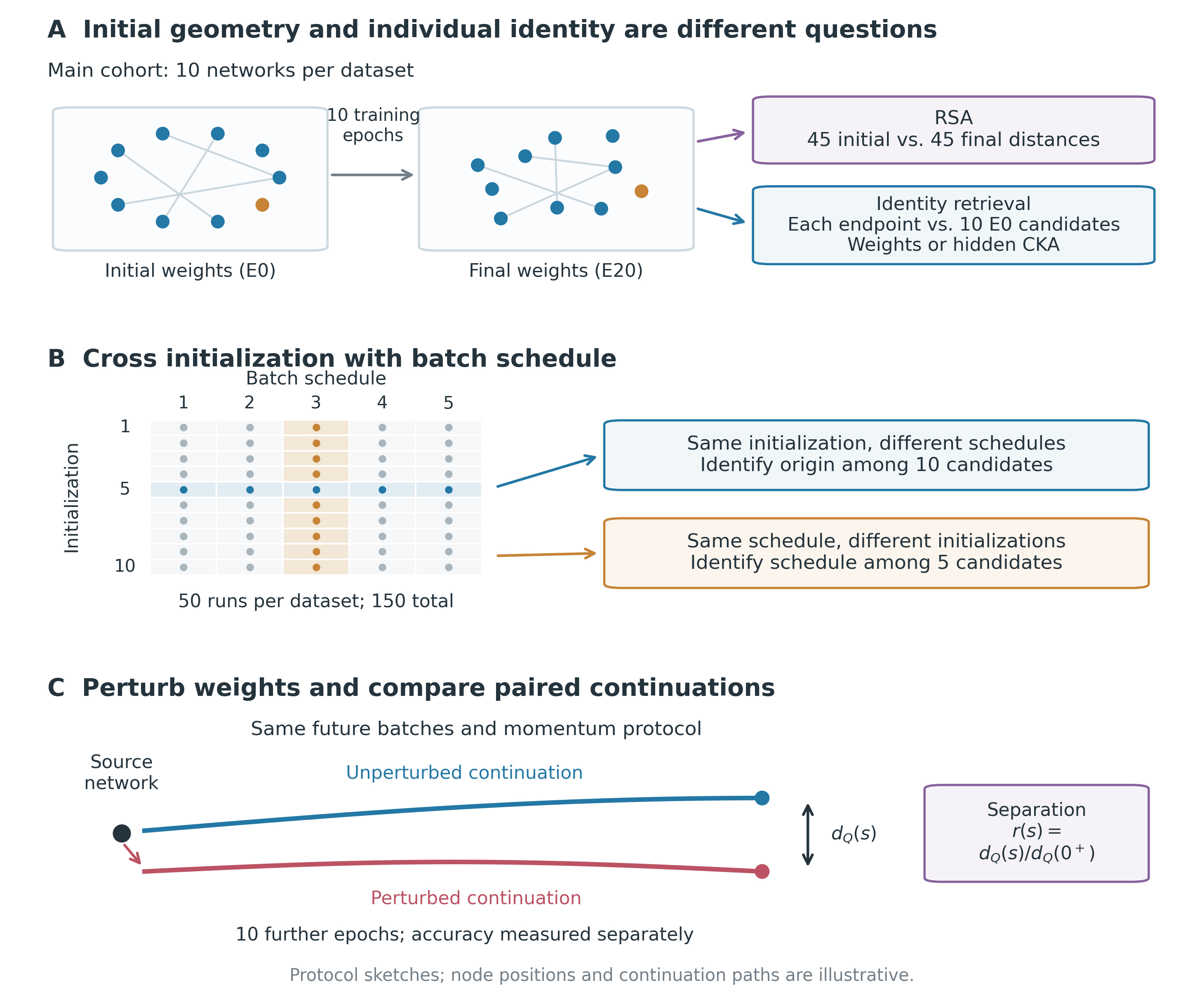}
\caption{\textbf{Experimental designs for investigating weight-space geometry, network identity, and learning trajectories.} (A) Ten independently initialized networks per dataset are compared in two ways. First, representational similarity analysis (RSA) compares the 45 pairwise distances between networks at initialization and after training to determine whether their original geometric arrangement is preserved. Second, each trained network is compared with all ten possible initializations to test whether its original starting point can still be identified. This identification is performed using either the network weights or the similarity of internal neural activations (CKA). (B) To distinguish the influence of initialization from that of training data order, ten initializations are combined with five different sequences of training batches. The highlighted rows and columns indicate networks that share either the same initialization or the same batch order. (C) To investigate how networks respond to disturbances during learning, a training trajectory is split into two branches. Random perturbations are added to the weights of one branch, while the other continues without perturbation. Both branches subsequently receive identical training batches and follow the same momentum protocol. We measure whether their weight configurations move closer together again, relative to their separation immediately after the perturbation, and whether task performance recovers. The initial MNIST experiment examines three perturbation strengths and five random directions per network. The extended experiment investigates three datasets, three intervention times, three layer conditions, four perturbation strengths, and two momentum protocols, using one random direction per network and condition. All illustrated positions and trajectories are schematic and do not represent measured results.}
\label{fig:comparison-design}
\end{figure}

Table~\ref{tab:design} provides an overview of the experimental designs, while Figure~\ref{fig:comparison-design} illustrates the main comparisons. The perturbation experiments comprise an initial MNIST study and a larger extension across all three datasets, which are reported separately because they differ in the number of perturbations per network. Further results and control analyses for all eight experiments are provided in Supplementary Sections~\suppref{supp:rsa}--\suppref{supp:trajectory}.

\begin{table}[tbp]
\centering\small
\caption{Experimental cohorts and the question addressed by each comparison. Descendants of one initialization or source network remain grouped during inference.}
\label{tab:design}
\begin{tabularx}{\linewidth}{@{}>{\raggedright\arraybackslash}p{.24\linewidth}>{\raggedright\arraybackslash}p{.29\linewidth}>{\raggedright\arraybackslash}X@{}}
\toprule
Cohort & Design & Main comparison \\
\midrule
Main cohort & 3 datasets; 10 runs each & Initial--final geometry, weight identity, and representation identity \\
Crossed training & 10 initializations $\times$ 5 batch schedules per dataset; 150 runs & Shared initialization across schedules; shared schedule across initializations \\
Perturbation pilot & 10 MNIST sources; 300 perturbed continuations plus controls & Disturbed versus unperturbed continuation \\
Perturbation extension & 3 datasets $\times$ 3 times; 2,160 perturbations and 180 baselines & Generalization across intervention time, layer, strength, and momentum \\
Paired task history & 200 seeds $\times$ 2 histories $\times$ 2 momentum protocols; 800 runs & A--B--C versus B--A--C after ten C epochs \\
Trajectory and events & Two cohorts of 200 seeds $\times$ 2 histories; 800 runs & Earlier reduced states and one or four AB/BA events after five C epochs \\
\bottomrule
\end{tabularx}
\end{table}

\subsection{Symmetry-adjusted weights and ensemble geometry}

For hidden neuron $h$, we balanced the norms of its incoming vector $u_h$ and outgoing vector $v_h$ using the positive factor $a_h=\sqrt{\norm{v_h}/\norm{u_h}}$, replacing them by $a_hu_h$ and $v_h/a_h$. This canonical ReLU gauge preserves the network function. Both norms must be nonzero, as they were for the analyzed networks. A single permutation is then applied jointly to the columns of $W_0$ and rows of $W_1$. For this one-hidden-layer architecture, the Hungarian assignment gives the exact optimum of the joint weight-matching objective used by Git Re-Basin \citep{ainsworth2023rebasin}.

The primary ensemble analysis correlated the upper triangles of the initial and current distance matrices within each dataset. Each matrix contains 45 unique network pairs. Following the logic of representational similarity analysis \citep{kriegeskorte2008rsa}, Spearman correlation measured preservation of the ordering of pairwise distances. The primary distance was the joint Frobenius distance after canonicalization and optimal matching. A complementary analysis normalized the complete canonical weight vector to unit length. Ensemble geometry and retrieval therefore share the same symmetry treatment, while their primary scale conventions are distinguished explicitly.

\subsection{Initialization retrieval and controls}

For each network state, let $W_0^c$ and $W_1^c$ denote the gauge-balanced weights and let $N=(\norm{W_0^c}_F^2+\norm{W_1^c}_F^2)^{1/2}$. We used $\widehat W_\ell=W_\ell^c/N$ for both layers ($\ell=0,1$). In the minimization below, $P$ ranges over all $64\times64$ permutation matrices of the hidden units. If $\widehat W$ denotes this normalized pair, the cost between current network $i$ and initial candidate $j$ was
\begin{equation}
 d_{ij}(t)=\min_{P}\left[
 \norm{\widehat W_{0,i}(t)-\widehat W_{0,j}(0)P}_F^2+
 \norm{\widehat W_{1,i}(t)-P^\top\widehat W_{1,j}(0)}_F^2
 \right]^{1/2}.
 \label{eq:distance}
\end{equation}
Total normalization is an analysis operation, not a function-preserving symmetry. Every current network was compared with all ten initializations from its dataset using exactly the same calculation. Identity labels were consulted only after ranking candidates. Mean correct-candidate rank was the primary endpoint at E20; chance rank is 5.5 and chance top-1 accuracy is 10\%. Exact ties received average ranks and fractional top-1 credit.

Controls compared raw and canonical weights, native neuron coordinates and optimal assignments, normalized and unnormalized distances, and the contributions of each layer. Layer contributions retained the joint normalization and joint assignment. We also selected the least and most active hidden-unit quartiles using only the current network's responses to a fixed probe. Small and large residual controls evaluated the lower and upper quartiles of the matched, normalized entrywise differences separately for each candidate pair. These are alternative costs, not interventions on historically unchanged weights. Actual initial-to-final raw-weight changes were measured separately. Norm-only controls were supplemented by a post hoc comparison of within-ensemble norm ranks, because common norm growth can make absolute norm differences rank all candidates identically.

\subsection{Representation comparisons and crossed training}

The main representation comparisons used a fixed, ordered probe of 3,000 test images, with 1,000 from each dataset. For feature matrices $X$ and $Y$, we computed centered linear CKA,
\begin{equation}
 \CKA(X,Y)=\frac{\norm{X_c^\top Y_c}_F^2}
 {\norm{X_c^\top X_c}_F\,\norm{Y_c^\top Y_c}_F},
\end{equation}
where columns are centered across probe inputs \citep{kornblith2019cka}. Retrieval used $1-\CKA$, without neuron matching. Raw hidden activations were primary. Feature-standardized hidden activations and raw logits provided complementary analyses; constant features were set to zero after centering. Standardization removes positive featurewise scale differences that raw CKA does not remove.

The crossed experiment combined each of the ten original initializations per dataset with five independently specified batch schedules. The initial weights were bitwise identical to the original E0 states. Each schedule specified the same sequence of training indices across initializations. All 150 networks trained for ten epochs. For initialization identification, an endpoint under one schedule was ranked against the ten endpoints under another schedule. The correct candidate shared its initialization. All 20 directed schedule pairings were averaged first within initialization and then across initializations. The primary comparison used Equation~\ref{eq:distance}; final hidden CKA was secondary. The reverse analysis searched for the same schedule among five endpoints from another initialization, with chance rank 3 and top-1 accuracy 20\%. Normalized rank scores, $(n-\bar r)/(n-1)$, place both candidate-set sizes on a common chance value of 0.5. Initial-to-final weight retrieval was also repeated across all schedules.

\subsection{Perturbation and continuation}

Ten trained MNIST networks at E20 were perturbed in both weight matrices by independent Gaussian directions normalized to 1\%, 5\%, or 20\% of the respective matrix norm. Five directions per strength were tested with momentum retained and with momentum reset, giving 300 perturbed continuations. For each source network and momentum protocol, the 15 perturbed continuations (three strengths and five directions per strength) shared one unperturbed control with the same future batch sequence, giving 20 controls in total. Both arms trained for ten more epochs. Separate controls changed the future batch sequence without a weight intervention.

The completed extension crossed all three datasets with E2, E10, and E20 (one, five, and ten training epochs), intervention on $W_0$, $W_1$, or both, strengths of 1\%, 5\%, 10\%, or 20\%, and both momentum protocols. Ten source networks per dataset and time each contributed one direction per cell, yielding 2,160 perturbed continuations and 180 matched baselines. All 2,340 continuations completed ten further epochs without numerical failure. The source trajectories are reused across intervention times; cells are not independent cohorts. Ninety independent reference continuations supplied time-matched comparison ensembles, including 20 newly trained reference initializations for Fashion-MNIST and KMNIST.

Let $d_Q(s)$ be the canonical, optimally matched joint Frobenius distance between perturbed and unperturbed networks after $s$ further updates. Unlike retrieval, this analysis retains total weight scale. We assessed contraction using
\begin{equation}
 r(s)=d_Q(s)/d_Q(0^+).
\end{equation}
Here $0^+$ is the instant immediately after the weight perturbation and before the next SGD update. A value below one indicates reduced separation from the paired continuation. Accuracy and loss were tracked independently of this structural criterion.

Ten independently initialized reference networks defined a task-compatible region using a structural distance envelope and performance thresholds. We distinguished a time-matched envelope from a fixed final-horizon region. A perturbation-induced return required membership before intervention, exit immediately after perturbation, and membership again at the final observation. This stricter fixed-region analysis was specified during pilot training, before inspection of pilot outcomes, as a secondary addition to the original plan. We also evaluated loss along eleven equally spaced points between aligned endpoints to separate endpoint connectivity from return dynamics.

For the extension, the fixed reference horizons were 11, 15, and 20 total training epochs. To test sensitivity to the assignment, we matched hidden units in two further ways. The first used Hungarian assignment on distances between permutation-invariant features (PIFs) summarizing each unit's incoming and outgoing weights. The second used Alpha profile costs, which weight sign disagreements by weight magnitude \citep{krauss2026clustering}. After either assignment we measured the same canonical joint Frobenius distance. The separate Alpha sign-dissimilarity measure was analyzed as an end-minus-start change, not a distance ratio. Supplementary Section~\suppref{supp:perturbation} provides region definitions, batch-order controls, matching comparisons, and the complete 216-cell grid. Cellwise bootstrap intervals are descriptive and are not adjusted for simultaneous comparisons across the grid.

\subsection{Task-history decoding}

\begin{figure}[tbp]
\centering\includegraphics[width=\linewidth]{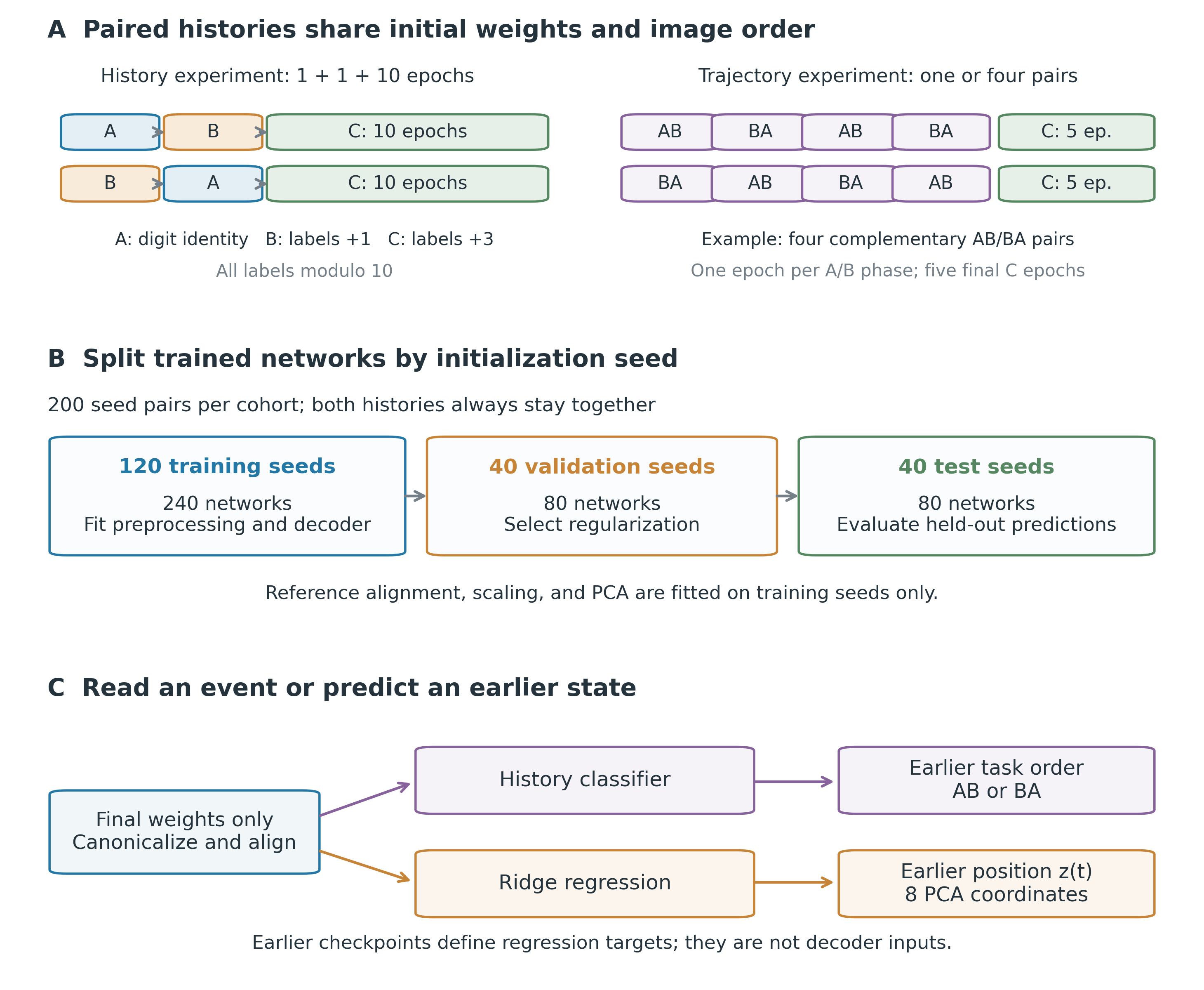}
\caption{\textbf{Decoding learning history from final network weights.} (A) Controlled learning histories. Two networks start with identical weights and encounter the same images, but learn two different tasks, A and B, in opposite order (A--B or B--A). Both are subsequently trained on the same final task C for ten epochs. Separate experiments investigating trajectory reconstruction use one or four successive A/B task pairs, followed by five epochs of C. In the four-pair experiment, task order is randomly chosen for each pair and reversed in the corresponding comparison network. (B) Independent evaluation. Networks sharing the same initialization are always assigned to the same training, validation, or test set. Consequently, the decoders are evaluated on previously unseen initializations. All preprocessing is fitted using training data only, while validation data are used to select regularization parameters. (C) Decoding from final weights. Linear probes predict either the order of previously learned tasks or earlier positions along the learning trajectory. Task-order classification uses logistic regression in the ten-epoch-C experiment and ridge classifiers in the trajectory experiments. Earlier positions are represented by eight PCA coordinates and predicted using ridge regression. The PCA is fitted on training data, and the neuron alignment determined from each final checkpoint is applied consistently to its earlier checkpoints. Importantly, probes receive only the final weights as input; earlier checkpoints are used exclusively to define the prediction targets.}
\label{fig:decoding-design}
\end{figure}

Figure~\ref{fig:decoding-design} outlines the paired curricula and endpoint-probe pipeline. To manipulate learning history independently of starting weights, we trained paired copies of 200 MNIST initializations on A--B--C and B--A--C. A used digit identity, B used $(y+1)\bmod 10$, and C used $(y+3)\bmod 10$. A and B each lasted one epoch, followed by ten C epochs. Both arms received the same images and temporal batch plan. The main cohort reset momentum at both phase transitions; a paired robustness cohort retained it. Together these comprised 800 runs.

The 200 seed pairs were split into 120 training, 40 validation, and 40 test blocks. Both histories of a seed stayed in the same split. A linear L2-logistic probe classified history from canonical endpoint weights aligned to a frozen training-only reference. Regularization was selected on validation blocks. The primary endpoint used both matrices after ten C epochs. Earlier C horizons, individual layer contributions after joint preprocessing, final logits, diagnostic features, and weights residualized against diagnostics were secondary. Diagnostics comprised accuracy, loss, two raw matrix norms, canonical hidden-activation RMS, and logit RMS. Logits and activation diagnostics used a fixed, approximately class-balanced probe of 256 MNIST test images. Residualization was a training-only linear regression. Neither earlier weights nor optimizer state, seed labels, or archive identifiers were decoder inputs.

\subsection{Trajectory and event probes}

A separate experiment trained two MNIST cohorts, each with 200 seed pairs and the same 120/40/40 split structure. The one-bit cohort used AB or BA followed by five C epochs. The four-bit cohort used four consecutive AB/BA pairs followed by five C epochs, with two complementary four-bit histories per seed. Each A/B phase lasted one epoch, and momentum was reset at each phase transition. We saved half-epoch checkpoints. An exploratory analysis of the original 30 runs was used for pipeline development and is not treated as independent confirmatory evidence.

For each model, a permutation inferred solely from its final canonical weights aligned the entire trajectory to a training reference. Layer scaling and a PCA basis were estimated from training trajectories. Ridge regression predicted earlier absolute coordinates $z(t)$ from final weights; eight target coordinates and the initial state were primary. We also predicted $z(t)-z(T)$, used 16 or 32 coordinates, separated layers, and fitted a small multilayer perceptron. Baselines included the training mean at time $t$, persistence of the known endpoint, and the displacement predictor $\overline z_{\rm train}(t)-z(T)$. The last control has exactly the mean-baseline absolute MSE and isolates the contribution of knowing the endpoint. The endpoint itself was excluded from reconstruction scores.

Linear ridge classifiers decoded each AB/BA position from final weights, with the earliest position primary. Entire complementary history vectors were swapped within seeds for event permutations. A final descriptive audit examined C performance and hidden-unit activity on 1,024 fixed test images. It retained every run and did not alter the curriculum, splits, or decoder choices.

\subsection{Inference and reproducibility}

Inference preserved the network-level dependence of the data. Ensemble correlations and ten-candidate retrieval used 10,000 identity permutations per dataset. Crossed analyses permuted factor identities consistently within blocks. The three primary dataset endpoints were Holm-adjusted separately within each corresponding analysis. Retrieval uncertainty used paired deletion of a network's query row and candidate column; crossed analyses deleted whole factor levels. Reduced candidate sets were calibrated before reporting intervals on the full scale. Perturbation intervals resampled the ten source networks after averaging directions within each source.

History and event probes used 999 seed-block label permutations, refitting the decoder and reselecting regularization. Secondary history tests and event feature-by-position tests were adjusted within their specified Holm families. Trajectory permutations instead reassigned complete held-out target trajectories between seed blocks with the fitted probe fixed; they test individual correspondence, not superiority to a mean predictor. Decoder intervals resampled held-out seed pairs. Perfect empirical scores can yield zero-width bootstrap intervals, so sample accuracy and population certainty are not equated.

Analysis decisions were recorded in local frozen plans and documented amendments. All result figures and supplementary tables were generated from archived numerical results; the two method figures are protocol schematics. Source hashes and generation code are retained in the full analysis archive. Existing source experiments are unchanged. This manuscript integrates the completed analyses; it does not add new model training or redefine their primary endpoints.

\section{Results}

\subsection{Training changes weight-space geometry without erasing network identity}

Training reorganized the relative positions of neural networks in weight space, but did not erase the information about their individual starting points.

\begin{figure}[tbp]
\centering\includegraphics[width=\linewidth]{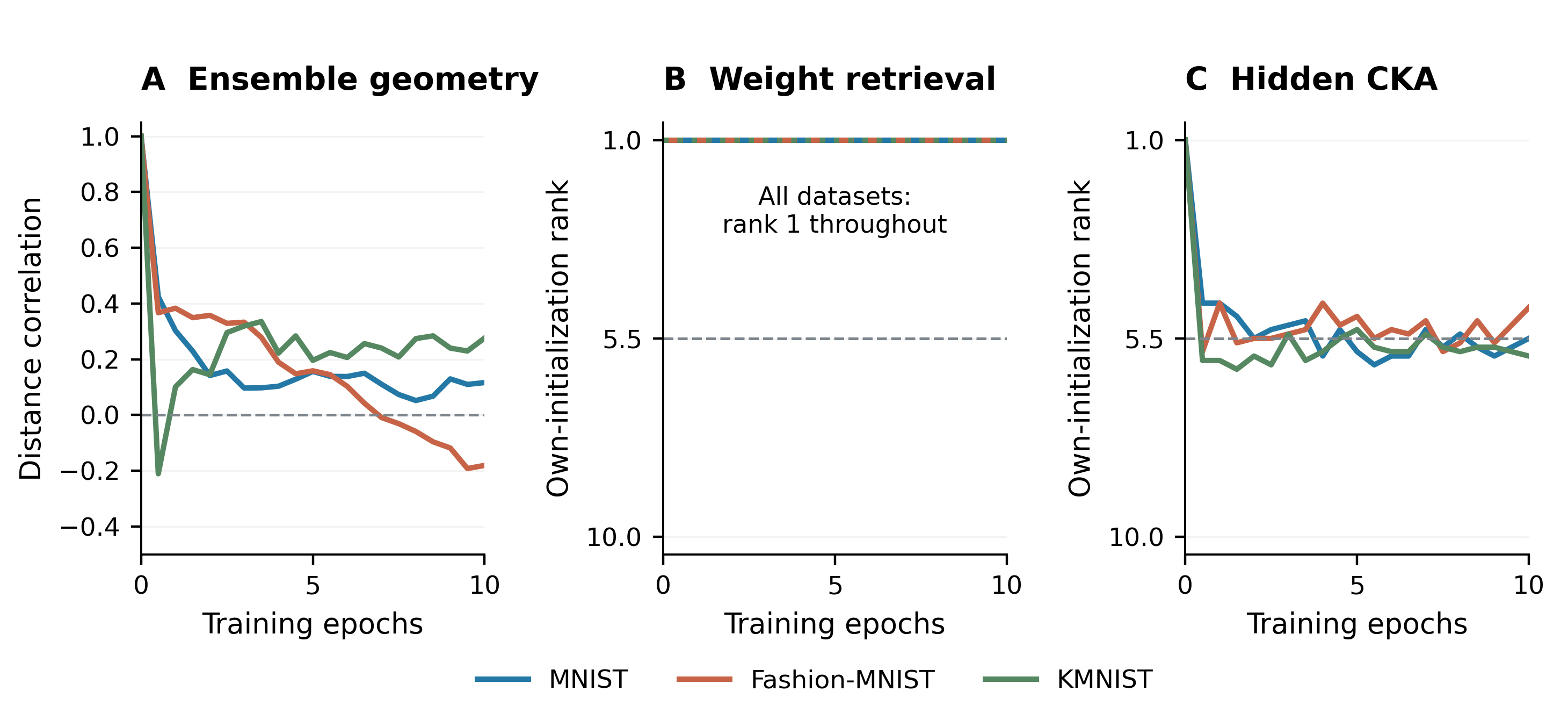}
\caption{\textbf{Training changes weight-space geometry while preserving individual network identity.} (A) Similarity between the initial and subsequent geometry of the network ensemble, measured by correlating pairwise distances in weight space (Spearman correlation, Frobenius distances after optimal neuron alignment). (B) Identification of each network's original initialization from its weight configuration during training. The correct initialization consistently ranks first among ten candidates for all three datasets (overlapping curves at rank 1). Weight matrices were jointly normalized and optimally aligned before comparison. (C) The same identification test based on internal neural activations rather than weights, using centered kernel alignment (CKA). Unlike weight-based identification, activation-based identification approaches chance level during training. Horizontal dashed lines indicate zero correlation in (A) and the chance-level mean rank of 5.5 in (B) and (C). Each dataset contains ten independently initialized networks evaluated at 21 checkpoints (E0--E20). E0 represents the initial self-comparison, and curves show observed values.}
\label{fig:identity}
\end{figure}

We first examined whether the geometry of randomly initialized networks was preserved during training. To this end, we compared the pairwise distances between networks before and after training. If the original geometry were preserved, networks that started close together should also end up relatively close together. However, this was not consistently the case. Correlations between initial and final distances were weak: 0.12 for MNIST, $-0.18$ for Fashion-MNIST, and 0.28 for KMNIST. None remained significant after correction for multiple comparisons, and the aggregate correlation was only 0.073. Normalizing the weights to unit length did not change this conclusion (Figure~\ref{fig:identity}A). Some additional analyses of individual layers and alternative alignment procedures showed positive correlations, but these relationships were not consistently observed across the primary comparisons (Supplementary Section~\suppref{supp:rsa}).

Despite this change in the overall geometry, each network retained a recognizable signature of its own initialization. We tested whether the final weight configuration of a trained network could be matched to its original initialization among ten possible candidates from the same dataset. Remarkably, all 30 networks were correctly matched to their own starting points. This perfect identification was maintained throughout training, at every saved checkpoint (Figure~\ref{fig:identity}B), and was statistically significant after correction in all three datasets. Importantly, all comparisons used the same normalization and optimal neuron-alignment procedure, ensuring that identification did not depend on the original ordering of neurons.

Interestingly, this persistent identity in weight space was not mirrored by a corresponding similarity of internal representations. We repeated the identification analysis using centered kernel alignment (CKA) to compare hidden activations before and after training. While each network was correctly matched to itself at initialization, this correspondence was not maintained throughout learning. By the end of training, the original initialization could no longer be reliably identified from the similarity of its hidden activations (Figure~\ref{fig:identity}C). Thus, the information about a network's origin remained accessible in its final weights even though its internal representations had been substantially reorganized.

Together, these findings reveal three distinct aspects of learning. First, training changes the overall geometry of an ensemble in weight space: networks that start close together do not necessarily remain close together. Second, despite this reorganization, each network retains an identifiable signature of its own initialization in its final weights. Third, this preserved identity does not imply that the original internal representations remain recognizable. Learning can therefore substantially reorganize both weight-space geometry and internal representations while preserving information about the individual starting conditions in the final weight configurations.

\subsection{Network identity persists despite extensive changes to the weights}

The ability to identify a network's original initialization could not be explained simply by differences in overall weight magnitude or by parameters that remained unchanged during training.

\begin{figure}[tbp]
\centering\includegraphics[width=\linewidth]{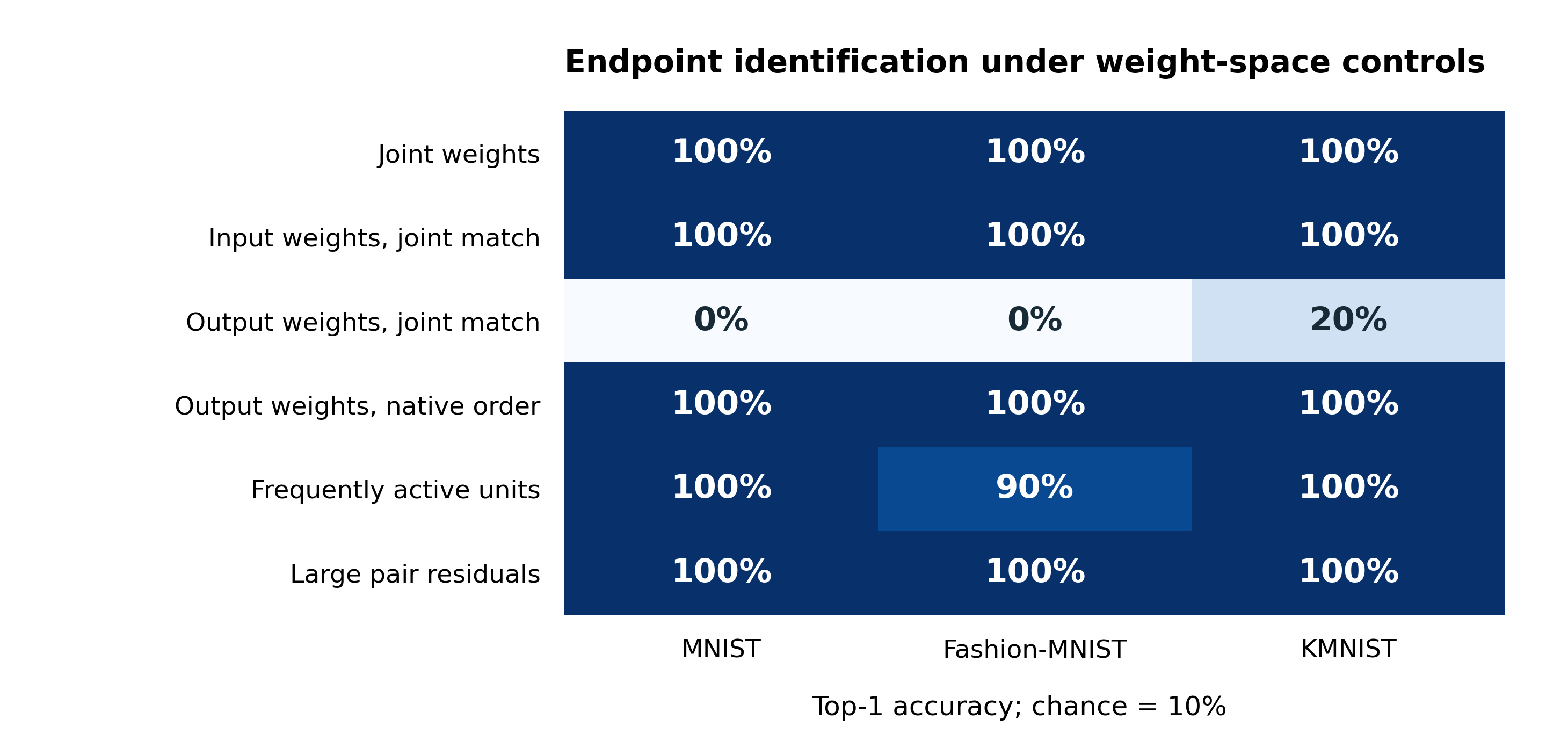}
\caption{\textbf{Identifying a network's initialization using different subsets of its final weights.} The figure shows how reliably each trained network can be matched to its original initialization among ten possible candidates, using different parts of its weight matrices and different comparison methods. Identification accuracy is measured as the percentage of networks for which the correct initialization ranks first (top-1 accuracy; chance level: 10\%). We separately examine input-layer weights, output-layer weights, and weights belonging to frequently active hidden neurons. We also test identification using only the 25\% largest differences between the trained weights and each candidate initialization. Input- and output-layer comparisons use canonicalized weights and, where indicated, optimal neuron matching. Frequently active neurons are selected based exclusively on the trained network's activity, while the largest weight differences are determined separately for each candidate comparison. All analyses use the same ten networks per dataset as Figure~\ref{fig:identity}.}
\label{fig:controls}
\end{figure}

To investigate where this information was preserved, we repeated the identification analysis using different representations of the weight matrices and different subsets of network parameters (Figure~\ref{fig:controls}). Remarkably, identification remained perfect when using the complete weight matrices, regardless of whether neurons were compared in their original order or optimally aligned, and whether weights were normalized or left at their original scale.

Examining individual network layers revealed a more nuanced picture. The input-layer weights alone were sufficient to correctly identify every initialization after optimal neuron alignment. For the output layer, however, the result depended strongly on how neurons were matched. Using the same optimal alignment, identification rates dropped to 0\%, 0\%, and 20\% for MNIST, Fashion-MNIST, and KMNIST, respectively. Yet when output weights were compared in their original neuron order, identification was again perfect. Thus, historical information was not necessarily absent from the output layer; its detectability depended on how the weights were compared.

We next asked whether this persistent identity could be explained by neurons that had barely participated in learning or by weights that had remained close to their initial values. This was not the case. Restricting the analysis to frequently active hidden neurons still allowed correct identification in 100\%, 90\%, and 100\% of cases across the three datasets. Moreover, no hidden neuron was completely inactive across the fixed set of test inputs, and not a single saved weight remained exactly unchanged from its initial value. The changes were substantial: the average displacement of the weight matrices corresponded to approximately 3.9 to 5.1 times their initial layer norms. Nevertheless, the networks retained recognizable signatures of their individual origins.

Additional controls confirmed that identification depended on the detailed structure of the weights rather than their overall magnitude. Even when comparisons were restricted to the largest differences between matched weight matrices, every initialization was correctly identified. Because these differences were determined separately for each candidate initialization, however, they do not identify a fixed subset of weights that stores the learning history. In contrast, using only weight norms produced identification close to chance, with average ranks of 4.92 for raw norms and 5.85 for canonical norms, compared with a chance expectation of 5.5.

Together, these results show that individual network identity survives extensive changes to the weight matrices. The information about a network's origin is therefore not simply preserved in unchanged parameters or overall weight magnitude, but remains accessible through the structured relationships between its weights. Additional layer-specific and normalization controls are reported in Supplementary Section~\suppref{supp:controls}.

\subsection{Initialization shapes final representations despite changes in activation geometry}

Although training substantially changes a network's internal representations, its initialization still influences which representations eventually develop. We first compared the hidden activations of each network before and after training using centered kernel alignment (CKA). Unlike the comparison of weight matrices, this analysis could not reliably identify a trained network's original initialization. The correct initialization achieved average ranks of 5.5 for MNIST, 4.8 for Fashion-MNIST, and 5.9 for KMNIST among ten candidates, with an overall top-1 identification rate of only 6.7\% (Figure~\ref{fig:identity}C). Standardizing the hidden features or comparing network outputs (logits) did not substantially improve identification. Thus, the original activation patterns were no longer sufficiently similar to the final representations to identify their individual origins.

\begin{figure}[tbp]
\centering\includegraphics[width=\linewidth]{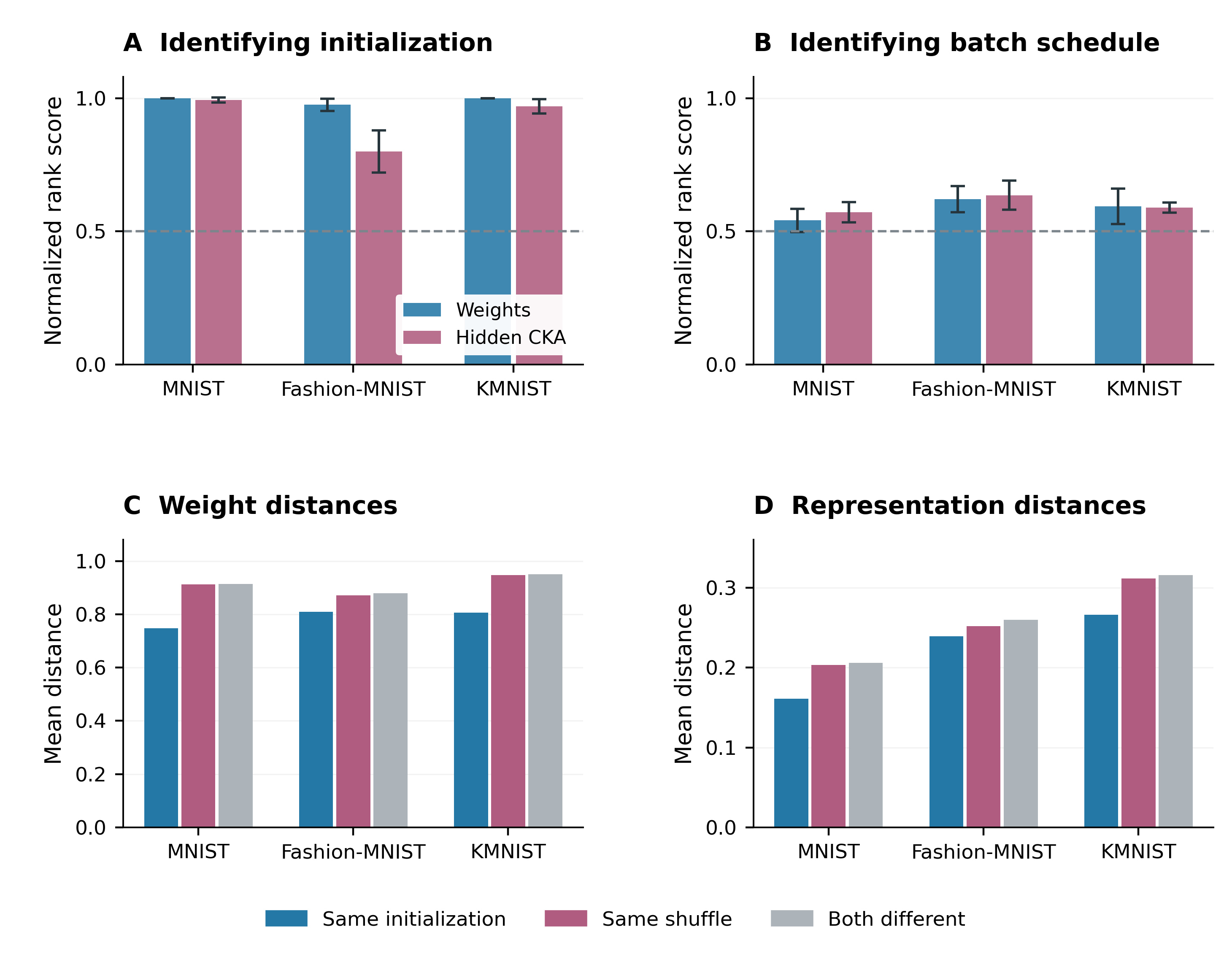}
\caption{\textbf{Both initialization and training batch order leave recognizable signatures in trained networks.} For each dataset, ten independently initialized networks were trained using five different sequences of training batches, resulting in 50 networks. (A) Identification of networks sharing the same initialization, despite being trained with different batch orders. (B) Identification of networks sharing the same batch order, despite starting from different initializations. Both analyses use either the final weight configurations or the similarity of internal representations measured by centered kernel alignment (CKA). Identification is expressed as a normalized rank score, where 0.5 represents chance performance and higher values indicate better identification. This normalization allows comparisons between the ten possible initializations in (A) and the five possible batch schedules in (B). Error bars indicate 95\% jackknife intervals that account for the grouping of networks by initialization and batch schedule. (C) Distances between final weight configurations and (D) differences between final internal representations, comparing networks that share the same initialization, share the same batch order, or differ in both. Because individual networks contribute to multiple pairwise comparisons, these distances are presented descriptively rather than treated as independent observations.}
\label{fig:crossed}
\end{figure}

However, this does not necessarily mean that initialization has lost its influence on the representations learned during training. To investigate this question, we trained additional networks using ten different initializations and five independently varied sequences of training batches. This allowed us to compare networks that started from identical weights but followed different learning trajectories. Remarkably, networks sharing the same initialization remained more similar after training, despite encountering the training batches in different orders. Using their final weights, networks with a shared initialization could be identified with accuracies of 100\% for MNIST, 90\% for Fashion-MNIST, and 100\% for KMNIST. Importantly, the same effect was also present in their internal representations, with identification rates of 97\%, 45.5\%, and 87\%, respectively (Figure~\ref{fig:crossed}A). Consistent with these findings, networks sharing an initialization were closer to one another in both weight space and representation space than networks differing in both initialization and batch order (Figure~\ref{fig:crossed}C,D).

These findings resolve the apparent contradiction. Although a network's final representations no longer directly resemble its initial activation patterns, the initial weight configuration still influences how the new representations are organized. Two networks starting from identical weights can therefore develop similar final representations, even when they follow different learning trajectories. Initialization leaves a recognizable influence on the geometry of the learned representations without preserving their original structure.

The order of training batches also left detectable traces in the final network configurations. Across datasets, the correct batch schedule could be identified from the final weights in 24\% to 33\% of cases, compared with a chance level of 20\% among five candidates (dataset-specific top-1 results in Supplementary Section~\suppref{supp:crossed}). Figure~\ref{fig:crossed}B shows the corresponding normalized mean ranks. Furthermore, all 150 trained networks could still be correctly matched to their individual initial weight configurations. Thus, both initialization and training batch order contribute information to the final network state, although initialization provides the stronger identification signal.

Importantly, the strength of these identification signals does not necessarily reflect their influence on task performance. For Fashion-MNIST, differences in batch order explained more of the variation in final classification accuracy than differences in initialization, even though initialization was much easier to identify from the final weights and representations. The complete analyses are provided in Supplementary Section~\suppref{supp:crossed}. These results demonstrate that a factor can leave a strong signature in the internal structure of a network without necessarily being the dominant determinant of its predictive performance.

\subsection{Networks recover performance without returning to their original learning trajectories}

Neural networks can recover their performance after a disturbance without returning to their original learning trajectories in weight space.

To investigate this, we introduced random perturbations into the weight matrices during training and compared each perturbed network with an unperturbed copy. Both networks subsequently received identical training batches. We then examined two questions independently: whether the perturbed network recovered its classification accuracy and whether its weights returned toward the unperturbed trajectory. The latter was measured by comparing the distance between both networks at the end of training with their distance immediately after the perturbation. A separation ratio above one indicates that the networks ended farther apart than they were immediately after the disturbance.

\begin{figure}[tbp]
\centering\includegraphics[width=\linewidth]{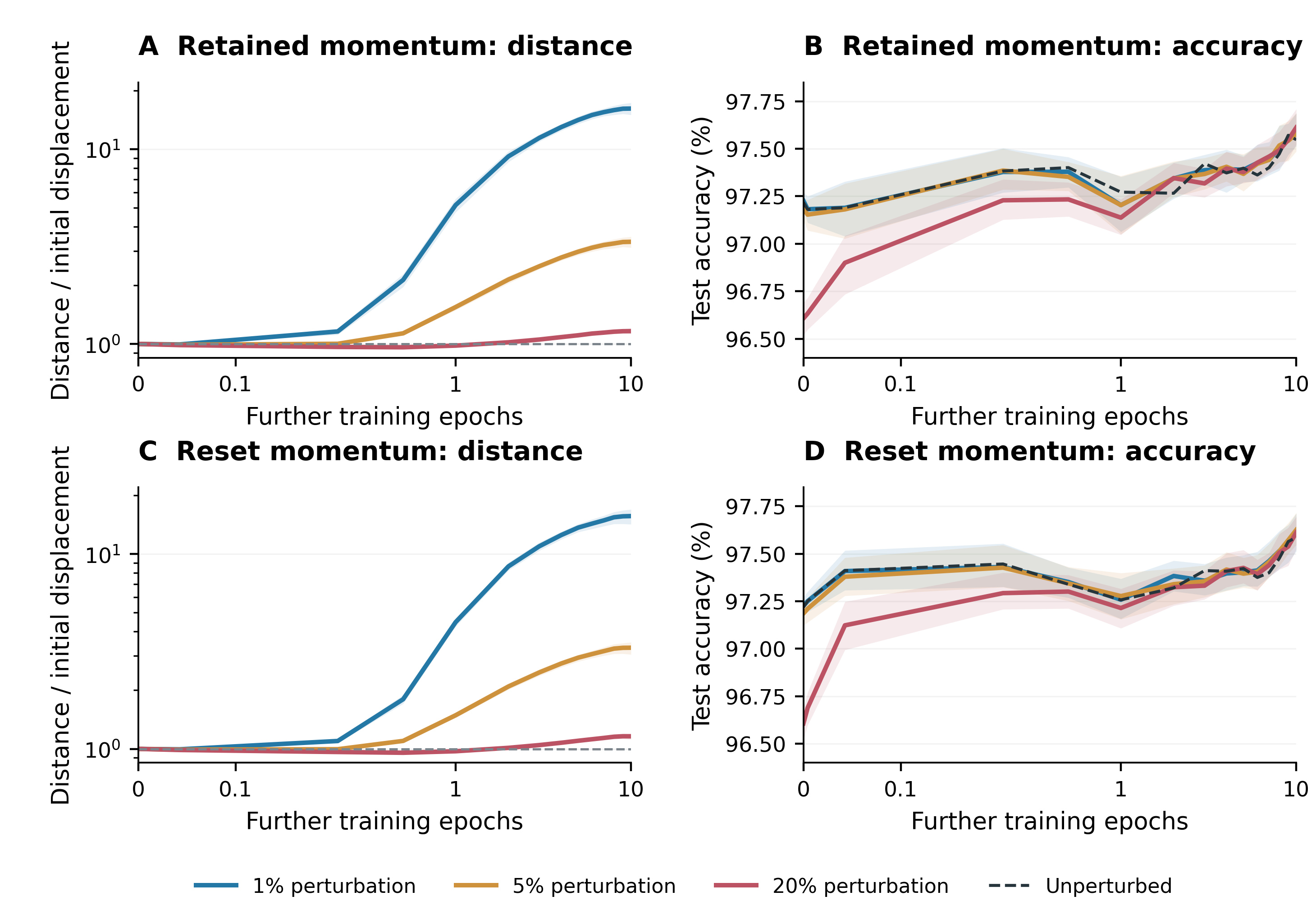}
\caption{\textbf{Networks recover their performance without returning to their original learning trajectories.} MNIST networks are perturbed at checkpoint E20 and then trained for another ten epochs alongside unperturbed copies. Both receive the same batches during continued training. (A,C) Distance between perturbed and unperturbed networks, normalized to their distance immediately after perturbation. Values below one indicate that the networks move closer together, while values above one indicate that they move farther apart. (B,D) Test accuracy of perturbed networks compared with their unperturbed counterparts. In the top row (A,B), momentum is retained; in the bottom row (C,D), momentum is reset in both networks. Shaded areas show 95\% bootstrap confidence intervals across ten initial networks, with five perturbation directions averaged per network. The time axis provides a more detailed view of the early training period after perturbation.}
\label{fig:perturbation}
\end{figure}

The initial MNIST experiment revealed that the networks did not consistently return to their original trajectories. All 300 perturbations produced at least a temporary reduction in the distance to the corresponding unperturbed network. However, this initial approach did not persist. After ten additional training epochs, the average separation ratio exceeded one under every combination of perturbation strength and momentum protocol, with lower confidence bounds also above one. When momentum was retained, the final ratios were approximately 16.1, 3.34, and 1.17 for perturbation strengths of 1\%, 5\%, and 20\%, respectively. Resetting momentum produced a similar pattern (Figure~\ref{fig:perturbation}). The particularly large ratio for the smallest perturbation reflects the very small initial separation used as the denominator, rather than a larger absolute distance between the final networks.

Despite this persistent separation in weight space, the perturbed networks recovered their classification performance. The strongest perturbation initially reduced average test accuracy from 97.22\% to 96.61\%. After continued training, accuracy increased to 97.61\%, almost identical to the 97.57\% achieved by the unperturbed networks. Thus, the networks recovered their function without returning to the same weight configurations or learning trajectories. Furthermore, linear interpolation between the aligned final weight configurations revealed no positive loss barrier at the sampled points. The final solutions were therefore connected by sampled low-loss paths, even though the training dynamics did not bring the networks back together.

\begin{figure}[tbp]
\centering\includegraphics[width=\linewidth]{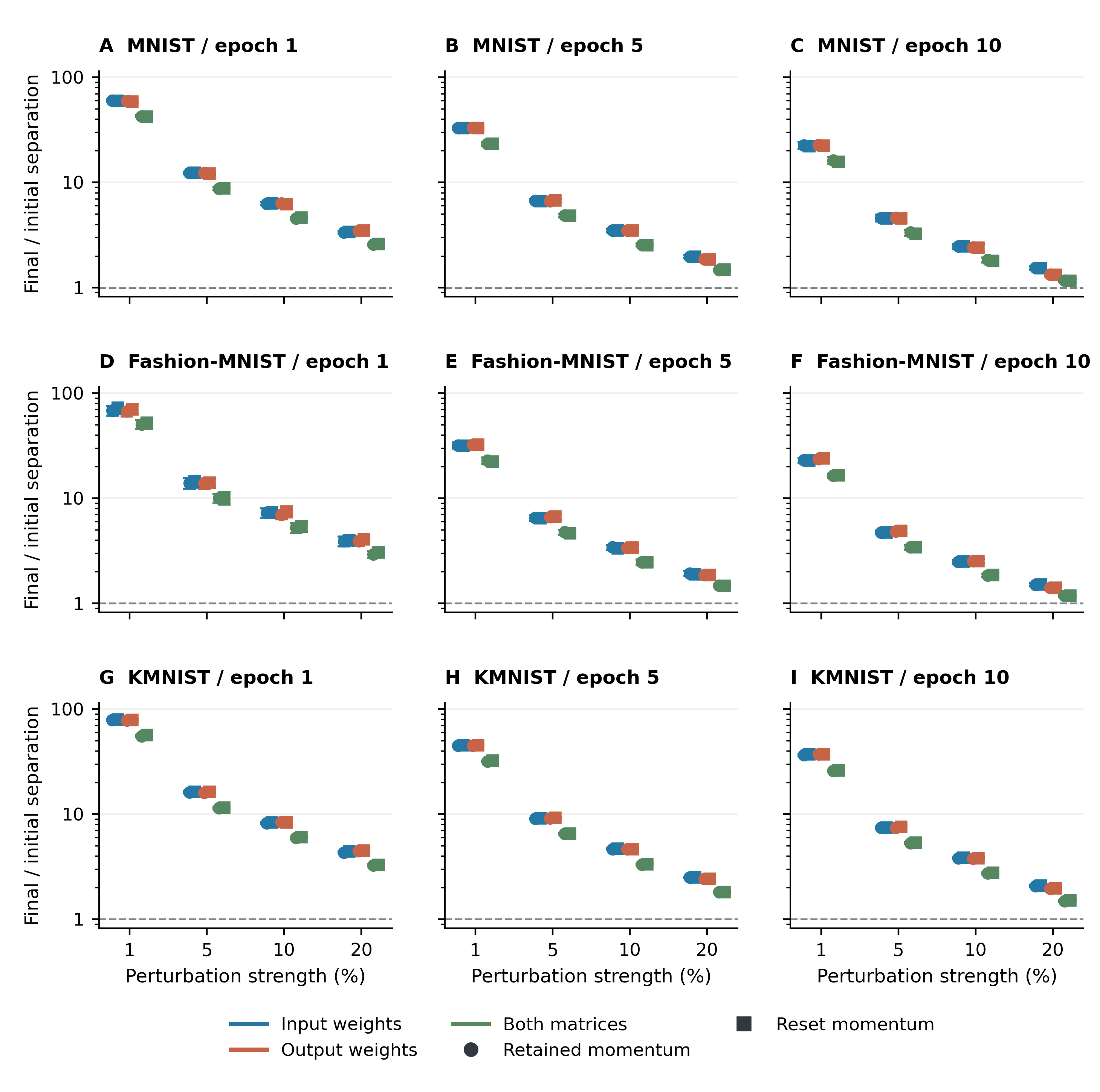}
\caption{\textbf{Perturbed networks move away from their original learning paths across datasets, training stages, and layers.} Each panel shows the results for one dataset and one perturbation time. We tested 24 conditions per panel, combining four perturbation strengths, three choices of perturbed layers, and two momentum settings (retained or reset). Points show the mean final distance between perturbed and unperturbed networks, relative to their distance immediately after perturbation. A value above one means that the networks have moved farther apart during continued training. Error bars indicate 95\% bootstrap confidence intervals across ten source networks. In all 216 conditions, even the lower confidence bound exceeds one (dashed line). Each condition includes ten networks, with one perturbation direction per network. The same source networks are used across conditions. The vertical axis uses a logarithmic scale. Confidence intervals are shown separately for each condition and are not corrected for multiple comparisons.}
\label{fig:expanded}
\end{figure}

To determine whether this finding was specific to MNIST or to particular perturbations, we extended the experiment to three datasets, three intervention times, three layer conditions, four perturbation strengths, and two momentum protocols. Across all 216 experimental conditions, the average final separation ratio remained above one, with its pointwise lower confidence bound also above one (Figure~\ref{fig:expanded}). This held whether perturbations were introduced after one, five, or ten training epochs, whether they affected one weight matrix or both, and whether momentum was retained or reset. Alternative methods for matching corresponding neurons confirmed the result: PIF-based matching \citep{krauss2026clustering} produced the same conclusion in all 216 conditions, while Alpha-based matching did so in 215 conditions, with one confidence interval including one. None of the alternative comparisons provided evidence of sustained convergence back toward the unperturbed trajectory.

The extended experiment also confirmed that structurally different final solutions can be connected through regions of low loss. Of the 2,160 pairs of perturbed and unperturbed endpoints, 2,157 showed no positive loss barrier along the sampled linear interpolation. The remaining three exhibited only small barriers, with a maximum increase of 0.00361 cross-entropy units. Thus, the existence of a low-loss connection between two solutions does not imply that training will dynamically drive one solution toward the other. Detailed separation ratios, performance measures, and alignment controls are provided in Supplementary Section~\suppref{supp:perturbation}.

These findings establish an important distinction between recovery of function and return to an individual learning trajectory. However, they do not yet establish whether networks are attracted toward a broader region of successful solutions in weight space. In the MNIST experiment, 180 of 300 perturbed networks eventually entered a predefined region satisfying both structural and performance criteria. Yet none of their unperturbed starting states had belonged to that same reference region, so these events cannot be interpreted as returns to a previously occupied region. In the extended experiment, all 2,160 networks remained inside the predefined structural region immediately after perturbation. The 115 subsequent returns to the combined structural-and-performance region therefore reflected recovery of the performance criterion rather than a departure from and return to the structural region.

Together, these findings reveal an extended set of functionally successful solutions connected by nearly barrier-free paths in weight space. Following perturbation, networks recover their performance without returning to their original learning trajectories. This behavior is consistent with attraction toward a shared set of solutions rather than a unique optimum, although direct evidence for attraction toward the geometrically defined low-loss region requires demonstrating recovery after leaving that region.

\subsection{The order of earlier learning remains detectable after a common final task}

We asked whether the final weights of a neural network still reveal the order in which it learned earlier tasks. To test this, we trained pairs of networks starting from identical weights and using the same training images in identical order, but with different task orders. One network learned A--B--C, while the other learned B--A--C. Both networks therefore experienced the same tasks, but in a different order. The final training phase on task C lasted ten epochs, ten times longer than either preceding phase. We then examined whether the weights still contained information about the earlier task order.

\begin{figure}[tbp]
\centering\includegraphics[width=\linewidth]{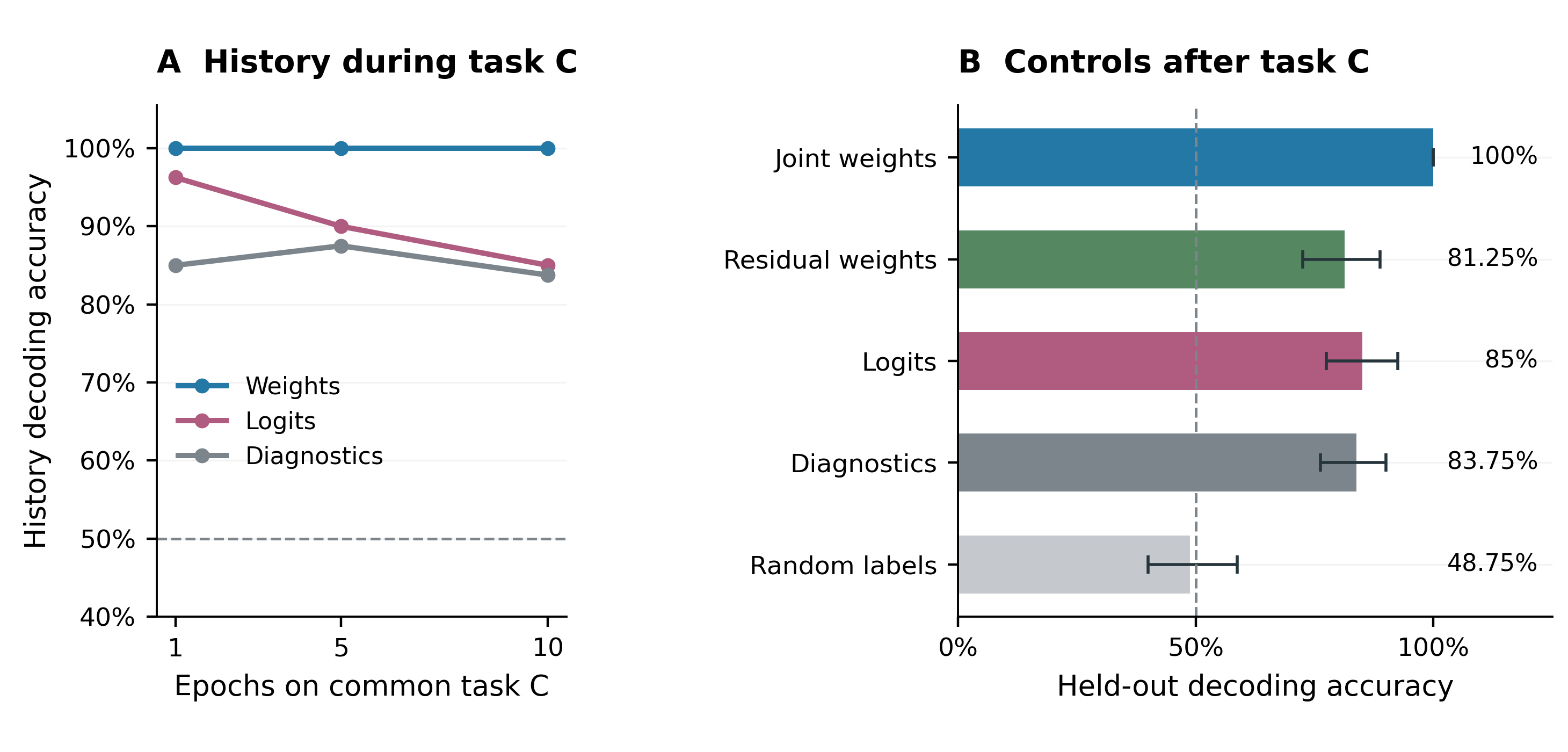}
\caption{\textbf{The order of earlier learning remains identifiable after training on a common final task.} (A) Accuracy of predicting whether networks learned task order A--B--C or B--A--C, measured after different amounts of training on task C. (B) Comparison of different decoding methods after ten epochs of task C. The residual-weight analysis tests whether task order can still be identified after removing the linear contributions of six measures related to performance, weight norms, and hidden activations. All results are evaluated on 80 networks from 40 previously unseen initializations. Both task histories belonging to the same initialization are always assigned to the same training or test set. Error bars show 95\% bootstrap confidence intervals across the held-out initialization pairs, using the selected decoder. Chance accuracy is 50\%. The 100\% weight-decoding accuracy produces a confidence interval with zero width in this sample, but does not imply perfect accuracy on all possible networks.}
\label{fig:history}
\end{figure}

The results were striking. After ten epochs of training on C, a linear decoder identified the earlier task order correctly for all 80 networks from 40 previously unseen initialization pairs, achieving 100\% accuracy. The same perfect accuracy was obtained after just one or five epochs of task C (Figure~\ref{fig:history}A). Importantly, both networks belonging to each test initialization were excluded from decoder training. The decoder therefore identified the learning history of networks with previously unseen initial weights, rather than simply recognizing individual training examples.

Several control analyses confirmed the robustness of this result. Resetting momentum between tasks did not affect decoding accuracy, and a separate experiment in which momentum was retained also achieved 100\%. When the history labels were randomly shuffled, accuracy dropped to 48.75\%, close to the chance level of 50\%. Both layer-specific weight analyses also achieved perfect identification at all three training durations. Additional controls confirmed that the aligned weight features were unchanged, to numerical precision, by neuron permutations and equivalent ReLU rescalings (Supplementary Section~\suppref{supp:history}).

Information about the earlier task order was also present in the networks' current outputs, although less reliably than in their weights. A decoder using the final logits achieved 85\% accuracy, while one based on performance measures and weight norms reached 83.75\%. Importantly, decoding from weights still achieved 81.25\% accuracy after removing the linear relationships with these diagnostic measures (Figure~\ref{fig:history}B). Thus, the historical information contained in the weights could not be fully explained by simple differences in performance or weight magnitude. Nevertheless, the two learning histories also affected final performance: mean accuracy on task C was 94.20\% for A--B--C and 93.39\% for B--A--C. The networks therefore did not produce identical functional outcomes.

Together, these findings demonstrate that prolonged training on the same final task does not necessarily erase information about earlier learning experiences. Even when networks start from identical weights, encounter the same tasks, and eventually optimize the same objective, the order of their earlier experiences remains detectable in their final weights. Crucially, this information can be decoded across previously unseen initializations, indicating a systematic influence of learning history rather than a signature specific to individual networks.

\subsection{Recognizing earlier learning experiences does not mean reconstructing the entire trajectory}

We next asked how much of a network's learning history could be recovered from its final weights. In particular, we wanted to distinguish two questions: Can we identify which tasks a network learned and in what order? And can we reconstruct the actual weights it had at earlier stages of training? To investigate this, we trained separate groups of networks on either one or four successive A/B task pairs, followed by five epochs of a common final task C. We then used the final weights to predict the order of earlier tasks and to estimate earlier weight configurations. For the latter, we represented the weights in a lower-dimensional space using principal component analysis (PCA) fitted only on the training data.

\begin{figure}[tbp]
\centering\includegraphics[width=\linewidth]{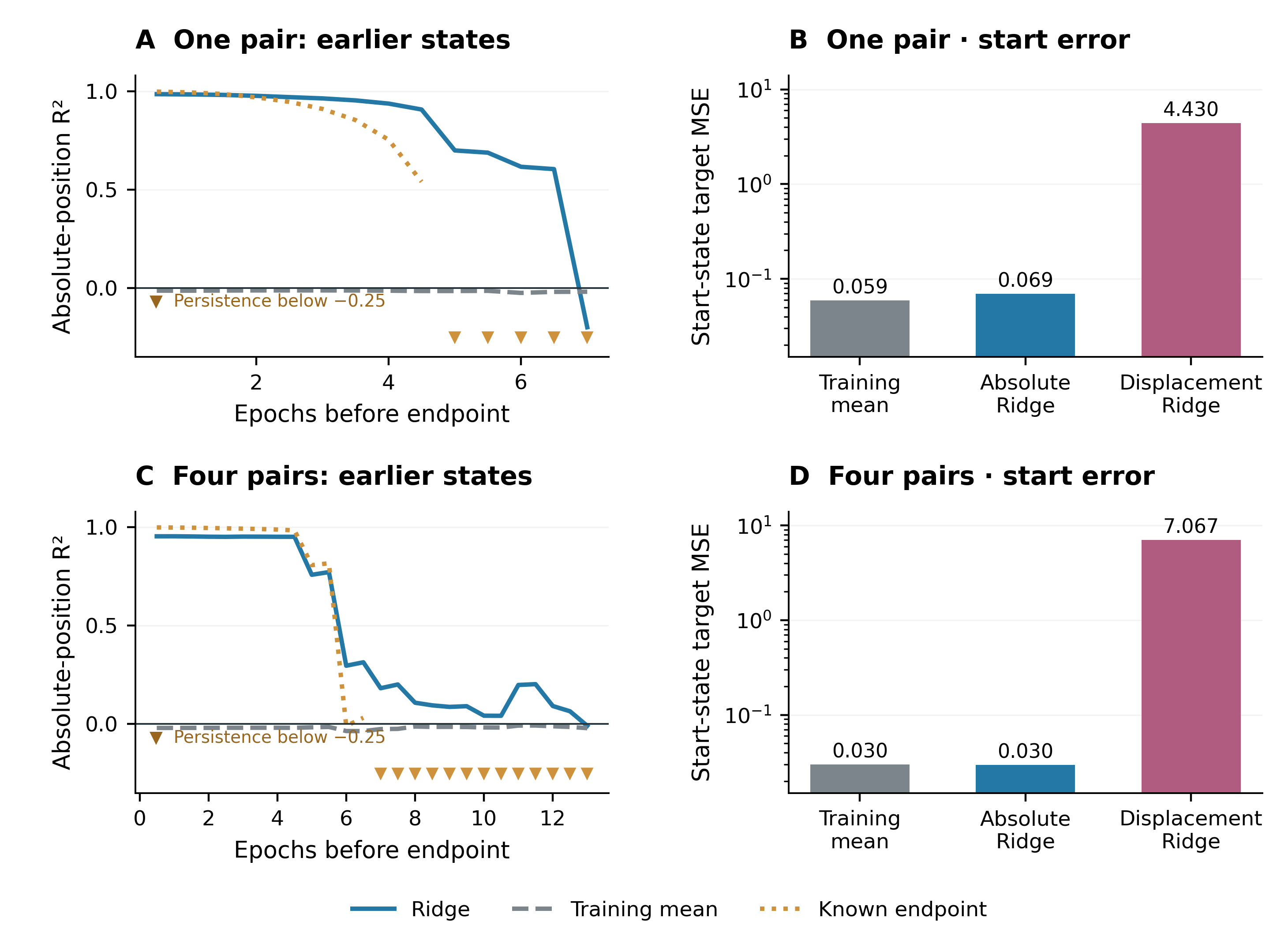}
\caption{\textbf{Some earlier weight states can be predicted, but the original initialization is difficult to reconstruct.} (A,C) Accuracy of predicting earlier weight configurations from the final weights, measured by $R^2$ at different times before the end of training. Ridge regression is compared with two simple baselines: predicting the average weights from the training data (training mean) or assuming that the earlier weights were identical to the final weights (persistence). Persistence scores below $-0.25$ are shown as downward triangles at this threshold. (B,D) Reconstruction errors for the original starting weights, comparing the training mean with Ridge models predicting either the earlier weights directly or their difference from the final weights. An additional displacement control is mathematically equivalent to the training-mean baseline and therefore produces the same error. Both experiments use eight PCA coordinates determined from training data and are evaluated on 40 held-out initialization pairs. The final checkpoint itself is not included in the reconstruction scores.}
\label{fig:trajectory}
\end{figure}

The results showed that some earlier weight configurations could indeed be estimated from the final weights. In networks trained on a single A/B pair, the predicted weights explained approximately 62\% of the variance after the first training epoch ($R^2\approx0.62$) and approximately 70\% after the two earlier task phases ($R^2\approx0.70$). These predictions were more accurate than simply using the average earlier weights or assuming that the weights had remained unchanged since the predicted checkpoint. However, this latter assumption became a strong baseline for checkpoints close to the end of training. Reconstructing the original initialization was considerably more difficult. Prediction accuracy was below the mean baseline for networks trained on one task pair ($R^2=-0.20$) and approximately equal to the mean baseline for networks trained on four pairs ($R^2\approx0$; Figure~\ref{fig:trajectory}). Thus, although the final weights retained information about certain intermediate stages of learning, they did not allow reliable reconstruction of the individual starting configurations.

At first glance, predicting the change in weights between initialization and the end of training appeared highly successful, with $R^2$ values above 0.95. However, a simple control achieved an even higher score of more than 0.999 by using only the average earlier weights and the known final weights. This revealed that the high scores were largely driven by information already contained in the final weights, rather than by an accurate reconstruction of the starting configuration. In other words, predicting the difference between two weight configurations is not necessarily the same as accurately predicting an unknown earlier configuration. Moreover, the eight PCA coordinates used in the main analysis captured only about 28\% of the variance in the scaled training trajectories.

We therefore tested whether more detailed representations or more flexible prediction methods could improve reconstruction. Increasing the number of PCA coordinates to 16 or 32, analyzing the network layers separately, or using a nonlinear decoder did not produce reliable reconstructions of individual starting weights. Additional tests with random prediction targets and shuffled labels served as controls for the analysis procedure (Supplementary Section~\suppref{supp:trajectory}). Thus, the final weights contained information about selected earlier states, but the tested methods could not reliably recover the original initialization.

\begin{figure}[tbp]
\centering\includegraphics[width=\linewidth]{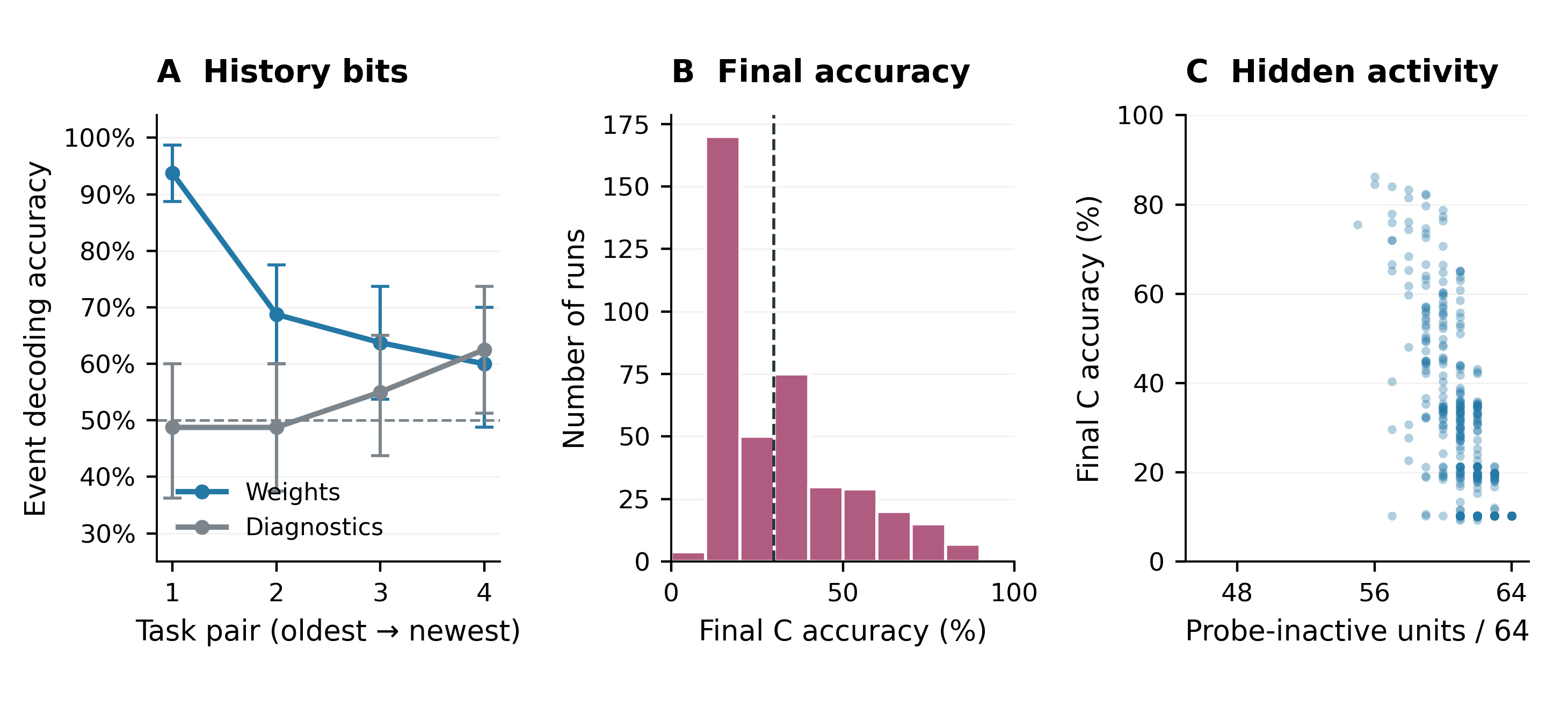}
\caption{\textbf{Earlier task orders remain identifiable even when networks perform poorly on the final task.} (A) Accuracy of identifying whether each of the four earlier task pairs was learned in the order A--B or B--A, using final weights or diagnostic measures. Error bars show bootstrap intervals obtained by resampling the 40 held-out initialization pairs. Weight-based decoding of the first two task pairs remains statistically significant after Holm correction for multiple comparisons. (B) Distribution of final accuracy on task C across all 400 networks in the four-pair experiment. The dashed line indicates the mean accuracy. (C) Relationship between final-task accuracy and the number of hidden neurons that remained inactive across 1,024 fixed diagnostic images. Panels B and C describe all networks in the experiment. No networks were excluded based on these measures, and the results do not establish that inactive neurons caused the poor learning performance.}
\label{fig:events}
\end{figure}

Identifying earlier learning events was considerably more successful. In networks trained on a single A/B task pair, the task order was again classified with 100\% accuracy. In the more complex experiment with four successive task pairs, the decoding accuracies were 93.75\%, 68.75\%, 63.75\%, and 60\%, from the earliest to the most recent pair. The first two results remained statistically significant after correction for multiple comparisons (Figure~\ref{fig:events}A). Interestingly, the earliest task pair was identified most reliably, rather than the most recent one.

However, learning history also influenced the networks' final performance. In the one-pair experiment, networks reached an average accuracy of 92.85\% on task C. The two learning histories differed in accuracy by approximately 1.09 percentage points across all 200 initialization pairs. Performance measures alone also contained information about task order. Thus, the historical differences were not restricted to the weights: they were accompanied by measurable differences in the networks' current behavior.

The four-pair experiment revealed an important limitation. After the extended sequence of earlier tasks, networks performed poorly on the final task C, reaching only 29.89\% accuracy on average, with individual results ranging from 9.24\% to 86.25\%. Their hidden-layer activity was also strongly reduced. On the fixed set of diagnostic images, an average of 61.16 out of 64 hidden neurons were inactive at the endpoint, and similarly low activity was already present before training on C (Figure~\ref{fig:events}B,C). All networks were retained in the analysis. Although earlier task orders remained decodable, these findings cannot be interpreted as evidence for a systematic temporal memory profile among equally successful networks. Instead, the extended experiment shows that traces of earlier learning can remain detectable even when subsequent learning is severely impaired.

\section{Discussion}

\subsection{Summary}

Learning the same task does not make neural networks historically identical. Our preceding study showed that networks trained on a common task develop related structures in permutation-aligned weights and internal representations \citep{krauss2026clustering}. The present results reveal what remains individual within this shared organization. Final weights retain information about a network's initialization and learning history, even after extensive training has changed its parameters. A task therefore constrains the solutions that learning can reach without uniquely determining how each solution is organized.

The distinction between shared structure and individual history appears at several levels. Training changed the relative arrangement of networks in weight space, yet each trained network remained identifiable by comparison with its own initialization. Networks starting from the same weights also developed recognizable final internal representations under different batch schedules, although their initial and final activation patterns were not directly matched by CKA. The influence of an initial condition can therefore survive learning without preserving its original form. Batch order contributed a smaller but detectable signature of its own.

Learning history was readable in a more direct sense when networks encountered the same tasks in different orders. Their final weights revealed whether tasks A or B had been learned first, including for initializations that the decoder had never seen. Yet identifying an earlier event was easier than reconstructing the weights of an earlier state. Some intermediate configurations could be predicted, whereas the original starting weights could not be recovered reliably with the tested models. A historical fingerprint is therefore not the same as a complete record of the trajectory.

The perturbation experiments add a further distinction. Networks recovered task performance after their weights were changed, but did not return to their unperturbed learning trajectories. Their different final configurations were nevertheless connected by almost barrier-free sampled interpolations. i.e. low-loss paths. Functional recovery, similarity of final weights, and return to a previous path are thus different properties. Together, these findings suggest a picture in which tasks impose a shared organization on weight space while leaving room for persistent differences associated with individual learning histories.

\subsection{Limitations}

Our experiments used small networks with one hidden layer, three image-classification datasets, and fixed-learning-rate SGD. These models made neuron matching and controlled comparisons feasible, but the same relationships have not yet been established for deeper networks, other optimizers, or more complex tasks. For example, rotations within self-attention can preserve a transformer's function \citep{zhang2025rotation}. Comparing its weights would therefore require accounting for symmetries beyond the hidden-unit permutations and positive rescalings used here. The main retrieval experiments used ten initializations per dataset. The crossed experiment added independently varied batch schedules, but did not increase the number of independent initialization identities. The history decoders were evaluated on held-out initializations within the tested task sequences; generalization to new curricula remains untested.

What can be recovered also depends on how networks are compared. Retrieval selected the correct origin from ten known candidates; it did not reconstruct an arbitrary, previously unknown initialization. Weight matching and normalization remove some irrelevant coordinate differences, but layer-specific comparisons depend on the joint alignment of the network. CKA measures similarity of activation geometry on a fixed set of inputs. Failure to identify an initial representation with this measure does not establish that all information about that representation has disappeared. One follow-up would repeat the same ten-candidate retrieval on the same probe images using projection-weighted canonical correlation analysis (PWCCA) \citep{morcos2018pwcca}, to test whether origin identification depends on the similarity measure. Similarly, the main trajectory reconstruction used eight PCA coordinates that captured only about 28\% of the scaled trajectory variance. Additional coordinates and alternative decoders did not solve the reconstruction problem, but these negative results do not demonstrate that recovery is impossible in principle.

The task-history experiments also require care in interpreting functional similarity. Although the networks learned the same final objective, their final accuracies and logits differed. The history signal remaining after diagnostic regression was not explained by the tested linear relationships, but other functional differences may still contribute. In the two-history comparison, task order was also linked to which task immediately preceded C. The complementary histories in the four-pair experiment changed several ordering choices together. Moreover, that longer curriculum produced poor final-task learning and widespread hidden-unit inactivity on the diagnostic inputs. Its early-event decoding result is valid for that cohort, but cannot establish a temporal memory profile among equally successful solutions.

The perturbation results are broad across datasets, intervention times, and layers, but do not establish attraction to a geometrically defined solution region. The extended experiment used one perturbation direction per source network and condition, compared with five in the pilot. Its 216 condition-wise confidence intervals are descriptive, share source trajectories, and were not adjusted for simultaneous coverage. Interpolation loss was evaluated at sampled points between aligned endpoints; these tests do not establish that an entire high-dimensional region has low loss. Crucially, the perturbed networks in the extension did not leave the predefined structural envelope immediately after intervention. Their subsequent recovery therefore does not demonstrate return to that region from outside. Constant learning rate and the finite follow-up also prevent conclusions about asymptotic convergence.

Finally, statistical independence lies with the source networks or seed pairs, not with every pairwise distance or checkpoint. The eight experiments were not conducted as one jointly preregistered family, and exploratory comparisons should not be treated as additional independent confirmation. Perfect identification in a finite held-out sample does not imply perfect performance in the population. These qualifications limit how broadly the present findings can be generalized without diminishing the central observation that specific aspects of learning history remain detectable.

\subsection{Related work}

The relationship between apparently different weight configurations has been investigated through neuron permutation, rebasin techniques, and mode connectivity \citep{ainsworth2023rebasin,garipov2018connectivity,krauss2026clustering}. Such studies show that solutions can appear distant because of equivalent parameterizations, and that different minima may be linked by low-loss paths. Our experiments build on these insights but ask a different question: after accounting for relevant symmetries, can the final weights still reveal how a particular network was trained? The answer is yes. Low-loss connectivity does not erase historical distinctions between the connected endpoints, nor does it imply that training moves back along a previously followed path. A related distinction between geometry and decodability arises when each image is encoded by the weights of an implicit neural representation. In that setting, stronger class clustering in weight space did not consistently improve a trained reader's classification accuracy \citep{guo2026routed}.

Studies using representational similarity analysis (RSA) and CKA compare the organization of neural activity across models and systems \citep{kriegeskorte2008rsa,kornblith2019cka}. The platonic representation hypothesis raises the possibility that different learning systems converge toward common representational structures \citep{huh2024platonic}. Our results complement this perspective by showing that shared task-related organization and individual historical signatures can coexist. Importantly, the final representations associated with a shared initialization need not resemble that initialization's original activation patterns. Earlier work on internal separability, successor representations, and learned construction representations likewise illustrates that different measurements reveal different aspects of what a network represents \citep{schilling2021separability,stoewer2022successor,ramezani2025constructions}. For example, successor representations can link visual and linguistic features in a common cognitive map \citep{stoewer2025multimodal}. In a next-word prediction network, hidden representations grouped by the grammatical class of the predicted word even without word-class labels during training \citep{surendra2023word}.

Other research has shown that initial conditions and early experience can shape later learning. The lottery ticket hypothesis links particular initial weights to successful optimization \citep{frankle2019lottery}, while work on critical learning periods shows that the timing of experience can have lasting consequences \citep{achille2019critical}. Continual-learning research usually asks whether performance on an earlier task is retained \citep{kirkpatrick2017forgetting}. However, historical decoding addresses a distinct question: can a later network state reveal what happened earlier, even without requiring the network to perform the earlier task? Our contribution is to test that question directly in final weights and to separate event identification from reconstruction of past states. In recurrent networks, \citet{herrmann2024rnn} inferred the training task from a trained model, with the best task identification obtained by probing its responses. Our paired curricula ask whether final weights distinguish two orders of the same earlier tasks after training on a common final task.

Underspecification research emphasizes that similar performance can arise from different learned predictors \citep{damour2022underspecification}. Our findings identify initialization, batch schedule, and task order as controlled sources of such diversity, while also showing that similar task performance should not automatically be called identical function. Twin-Merging uses a decomposition into shared and task-specific parameter components to combine separately fine-tuned models \citep{lu2024twin}. Our paired design asks which differences arise within models that reach the same final task by different routes. A related conceptual distinction appears in our work on recurrent computation, where useful processing depends on the interaction of network structure, input, and dynamics of evolving network states \citep{krauss2019recurrence,metzner2022dynamics,metzner2025reservoir,metzner2025organization}. The present study concerns a different dynamical system: it is the weights, rather than only the activations of a fixed network, that evolve during learning.

\subsection{Interpretation}

The findings strengthen the geometric basis of the algorithmic-attractor hypothesis proposed in our preceding study \citep{krauss2026clustering}. This hypothesis suggests that learning may be organized around task-dependent sets of successful solutions rather than a single preferred weight configuration. Our perturbation experiments now provide evidence that different endpoints can remain connected through low-loss paths. In 2,157 of 2,160 aligned endpoint pairs, the sampled linear interpolations showed no positive loss barrier. Networks can therefore recover their performance at different, nearly linearly connected weight configurations without returning to their original learning trajectories. These findings are consistent with an extended, set-based interpretation of algorithmic attractors. Whether such solution sets also exhibit the dynamical stability and restoring behavior required for genuine attraction remains an open question.

An important open question concerns the geometry of these connected low-loss solutions. The observed interpolation paths are compatible with two different structures: a broad, higher-dimensional region of low loss, or a much thinner, branching network of low-loss corridors separated by regions of higher loss. Both could support many functionally successful solutions, but they offer different possibilities for preserving learning history. In a broad solution region, historical information might be reflected in the position along weakly constrained dimensions. In a filament-like network, it could instead be associated with the particular branches or paths followed during learning. Both geometries could also support attraction toward a common solution set without requiring convergence to a single point. In particular, learning might be strongly constrained toward the solution structure while retaining freedom of movement within it. This offers a possible explanation for the coexistence of functional convergence and persistent historical individuality, although the geometry of these remaining degrees of freedom has yet to be established.

Our retrieval and history-decoding results show that final weight configurations retain information about their initialization and earlier learning experiences. These historical signatures persist despite substantial changes in the weights and remain detectable even in frequently active hidden units. They therefore cannot be explained simply by inactive neurons or unchanged parameters.

This view separates two forms of convergence. Networks may converge in what they can do without converging to the same weights or following the same trajectory. An extended solution set could therefore accommodate many historically distinct networks. It might also act as a dynamical attracting set: after displacement, learning could return a network to the broader set without returning it to its original point in weight space. That possibility is consistent with the observed functional recovery and endpoint separation. However, it remains distinct from the geometric connectivity already observed, because re-entry after leaving the structural region was not demonstrated.

The history experiments suggest a second, more general way to understand a learned model. Its present weights are not only a description of its current task competence; they can also be a partial record of the process that produced them. That record is selective. Knowing which of several initializations was used, or which task came first, is not the same as recovering the earlier weight values. A present state may therefore contain a compressed and incomplete representation of its own history. From these traces, a decoder may infer plausible earlier events or states without reconstructing the actual trajectory in full. The distinction between a historical trace and a faithful reconstruction is central: information about the past can survive even when the past is not directly recoverable.

The activation results reveal another important distinction. Although trained networks no longer reliably resembled their own initial internal representations, networks sharing the same initialization developed recognizable final internal representations across different batch schedules. Historical influence can therefore persist through transformation rather than preservation of the original activity patterns. The representation a network develops reflects not only the task it learns, but also the initial conditions under which learning began.

\subsection{Outlook}

A central question is where learning history is stored within the space of successful solutions. One possibility is that historical information occupies weight directions that have little influence on task performance. This can be tested by identifying directions according to their effect on task loss and selectively perturbing them. If weakly constrained directions carry historical information, perturbations along these directions should impair origin or history decoding more strongly than matched controls, while leaving task performance largely intact. Testing this relationship across different architectures, optimizers, and tasks would establish how general this organization is.

A second question concerns the geometry of the successful solutions themselves. Do they occupy broad, continuous regions of low loss, or are they connected by a thinner network of low-loss corridors? The present interpolation results cannot distinguish these possibilities. Future experiments could align several independently trained networks and measure loss not only along the lines connecting them, but also within the triangles and higher-dimensional simplices they form. \citet{sharma2024simultaneous} distinguish pairwise alignments from a simultaneous alignment and examine interpolation among three networks. For our proposed test, each endpoint would retain one neuron ordering across the whole simplex, so that measurements along its edges and in its interior refer to the same aligned weights. Low loss throughout these interiors would support extended solution regions, whereas low loss concentrated near connecting paths would favor a more filament-like structure. Perturbations perpendicular to these paths could reveal their local width. Combining these geometric measurements with historical decoding would help determine whether learning history is reflected in continuous variation within a solution region or in the particular branches and paths followed during training.

Once these structures are better characterized, their dynamical properties can be tested directly. Networks could be perturbed beyond a predefined low-loss region and then trained further while monitoring their distance from that region and their performance. A consistent return from outside would provide evidence for attraction toward an extended solution set, even if individual networks reached different endpoints. This would distinguish a geometrically connected family of solutions from a dynamically attracting one.

Finally, the limits of historical reconstruction deserve closer examination. Future experiments should use learning sequences in which earlier events vary independently while final-task performance remains comparable. Combining richer descriptions of network states with more powerful decoders could reveal how much of the past can be reconstructed from the present. Predictions of earlier states and trajectories should be evaluated against strong baselines in both reduced and full weight spaces. This would clarify whether the historical information retained by trained networks supports only the recognition of earlier events or also a more detailed reconstruction of their development.

The broader implication is that a trained network is not simply a solution to a task. It is one of many possible solutions, and its particular configuration can retain information about where learning began and how that solution was reached.

\FloatBarrier
\section*{Code and data availability}
The project repository is hosted at \url{https://github.com/krausspk/Learning-Trajectory}. The eight experiment directories contain analysis plans, code, numerical results, and provenance records. The scripts in \href{https://github.com/krausspk/Learning-Trajectory/tree/37b24bc504ffb39c491f28db5c6f18ef820422c2/manuscript/scripts}{\texttt{manuscript/scripts/}} generate the figures and supplementary tables from the CSV and JSON files in \href{https://github.com/krausspk/Learning-Trajectory/tree/37b24bc504ffb39c491f28db5c6f18ef820422c2/manuscript/data}{\texttt{manuscript/data/}}, including retrieval scores, perturbation outcomes, and history-decoding results. The \href{https://github.com/krausspk/Learning-Trajectory/blob/37b24bc504ffb39c491f28db5c6f18ef820422c2/README.md}{repository instructions} describe how to download the original image datasets and regenerate the model checkpoints and intermediate arrays locally.

\begingroup
\small
\setlength{\bibsep}{4pt}
\interlinepenalty=10000
\bibliographystyle{plainnat}
\bibliography{refs}
\endgroup

\clearpage
\pagenumbering{arabic}
\renewcommand{\thepage}{S\arabic{page}}
\setlength{\parskip}{2pt}
\setlength{\LTpre}{5pt}
\setlength{\LTpost}{5pt}
\pdfbookmark[0]{Supplementary material}{supplement}
\begin{center}
{\LARGE\bfseries Supplementary material\par}
\end{center}
\medskip

\appendix
\setcounter{section}{0}
\setcounter{table}{0}
\renewcommand{\thesection}{S\arabic{section}}
\renewcommand{\thetable}{S\arabic{table}}
\renewcommand{\theHsection}{supp.\arabic{section}}
\renewcommand{\theHtable}{supp.\arabic{table}}

This supplement reports the extended results and controls for experiments 01--08 in the order of their source analyses. Its tables are generated from archived numerical results retained in the full analysis archive. The main text presents the connected scientific argument; the supplement distinguishes primary endpoints, secondary comparisons, descriptive diagnostics, and the conditions under which each result was obtained. The same network, seed, or source-network block remains the inferential unit throughout. No additional training or model selection was performed for this manuscript revision.

\section{Ensemble geometry and its sensitivity to measurement (01)}
\label{supp:rsa}

The primary RSA compares the 45 dependent within-dataset pair distances among ten networks, with three dataset-level tests. It does not treat network pairs as independent observations. The initial--final correlations are 0.1165, $-0.1806$, and 0.2764; none survives the primary Holm correction. Across all 52 measurement variants, some layer-specific and fixed-assignment comparisons retain positive associations. For example, canonical output-layer Frobenius distances in Fashion-MNIST yield $\rho=0.4804$, and raw output-layer distances in KMNIST yield $\rho=0.4178$. These exploratory findings qualify a universal claim of geometric erasure. The supported result is that initial distances do not consistently predict final distances under the primary comparisons.

All Frobenius layer contributions use a joint assignment of both matrices. Balanced Frobenius distance is the square root of the mean of the two layer mean-squared differences, with 50,176 and 640 entries in $W_0$ and $W_1$, respectively. Cosine dissimilarity and unit-vector Frobenius distance are monotonic transforms under the same assignment and therefore produce identical Spearman ranks; they are not independent robustness tests. Alpha and SND comparisons retain the metric-specific neuron-profile assignments of the original reproduction and describe different geometries. Exact definitions are archived alongside the tables.

\begingroup\footnotesize\setlength{\tabcolsep}{4pt}
\renewcommand{\arraystretch}{1.12}
\begin{longtable}{@{}llllrrr@{}}
\caption{All 52 RSA endpoint variants. Each dataset entry is Spearman $\rho$ / unadjusted one-sided permutation $p$. Only the three canonical, optimally matched, joint Frobenius endpoints form the primary Holm family. All other rows are exploratory. $\alpha_2$ denotes Alpha sign dissimilarity with exponent 2. SND denotes Structural Network Dissimilarity. Size-wtd.\ weights the two layer values by their parameter counts (50,176 in $W_0$ and 640 in $W_1$).}\label{tab:s-rsa}\\
\toprule
Distance & Layer & Assignment & Gauge & MNIST & Fashion & KMNIST \\
\midrule\endfirsthead
\multicolumn{7}{l}{\textit{Table \thetable\ continued}}\\
\toprule
Distance & Layer & Assignment & Gauge & MNIST & Fashion & KMNIST \\
\midrule\endhead
\midrule\multicolumn{7}{r}{\textit{Continued on next page}}\\\endfoot
\bottomrule\endlastfoot
Frob. & Both & Opt. & Canon. & 0.116 / 0.2935 & -0.181 / 0.8401 & 0.276 / 0.0490 \\
Frob. & $W_0$ & Opt. & Canon. & 0.087 / 0.3641 & -0.410 / 0.9575 & 0.263 / 0.0682 \\
Frob. & $W_1$ & Opt. & Canon. & 0.054 / 0.3572 & 0.480 / 0.0007 & -0.002 / 0.5079 \\
Frob. & Balanced & Opt. & Canon. & 0.030 / 0.4137 & 0.468 / 0.0009 & -0.001 / 0.5049 \\
Cosine & Both & Opt. & Canon. & 0.052 / 0.3581 & 0.057 / 0.3396 & 0.130 / 0.2085 \\
Cosine & $W_0$ & Opt. & Canon. & 0.154 / 0.1421 & -0.073 / 0.6807 & -0.074 / 0.6873 \\
Cosine & $W_1$ & Opt. & Canon. & -0.030 / 0.5739 & 0.492 / 0.0007 & -0.064 / 0.6632 \\
Unit Frob. & Both & Opt. & Canon. & 0.052 / 0.3581 & 0.057 / 0.3396 & 0.130 / 0.2085 \\
Unit Frob. & $W_0$ & Opt. & Canon. & 0.154 / 0.1421 & -0.073 / 0.6807 & -0.074 / 0.6873 \\
Unit Frob. & $W_1$ & Opt. & Canon. & -0.030 / 0.5739 & 0.492 / 0.0007 & -0.064 / 0.6632 \\
Frob. & Both & Fixed E0 & Canon. & 0.299 / 0.0449 & 0.358 / 0.0128 & -0.029 / 0.5485 \\
Frob. & $W_0$ & Fixed E0 & Canon. & 0.291 / 0.0070 & 0.133 / 0.2045 & -0.264 / 0.9118 \\
Frob. & $W_1$ & Fixed E0 & Canon. & 0.209 / 0.0945 & 0.381 / 0.0055 & 0.199 / 0.1110 \\
Frob. & Balanced & Fixed E0 & Canon. & 0.215 / 0.0903 & 0.380 / 0.0054 & 0.191 / 0.1200 \\
Cosine & Both & Fixed E0 & Canon. & -0.042 / 0.6144 & 0.385 / 0.0054 & 0.157 / 0.1691 \\
Cosine & $W_0$ & Fixed E0 & Canon. & 0.076 / 0.3067 & -0.235 / 0.9403 & -0.047 / 0.6263 \\
Cosine & $W_1$ & Fixed E0 & Canon. & 0.056 / 0.3502 & 0.404 / 0.0037 & 0.239 / 0.0654 \\
Unit Frob. & Both & Fixed E0 & Canon. & -0.042 / 0.6144 & 0.385 / 0.0054 & 0.157 / 0.1691 \\
Unit Frob. & $W_0$ & Fixed E0 & Canon. & 0.076 / 0.3067 & -0.235 / 0.9403 & -0.047 / 0.6263 \\
Unit Frob. & $W_1$ & Fixed E0 & Canon. & 0.056 / 0.3502 & 0.404 / 0.0037 & 0.239 / 0.0654 \\
Frob. & Both & Fixed E20 & Canon. & 0.156 / 0.2395 & 0.166 / 0.1727 & 0.320 / 0.0167 \\
Frob. & $W_0$ & Fixed E20 & Canon. & 0.052 / 0.4146 & -0.359 / 0.9270 & 0.243 / 0.0938 \\
Frob. & $W_1$ & Fixed E20 & Canon. & 0.077 / 0.3167 & 0.424 / 0.0073 & 0.084 / 0.2867 \\
Frob. & Balanced & Fixed E20 & Canon. & 0.080 / 0.3155 & 0.420 / 0.0074 & 0.081 / 0.2927 \\
Cosine & Both & Fixed E20 & Canon. & 0.144 / 0.2337 & 0.245 / 0.0813 & 0.109 / 0.2396 \\
Cosine & $W_0$ & Fixed E20 & Canon. & 0.089 / 0.3015 & -0.074 / 0.6633 & 0.070 / 0.3307 \\
Cosine & $W_1$ & Fixed E20 & Canon. & 0.098 / 0.2649 & 0.436 / 0.0032 & 0.031 / 0.4160 \\
Unit Frob. & Both & Fixed E20 & Canon. & 0.144 / 0.2337 & 0.245 / 0.0813 & 0.109 / 0.2396 \\
Unit Frob. & $W_0$ & Fixed E20 & Canon. & 0.089 / 0.3015 & -0.074 / 0.6633 & 0.070 / 0.3307 \\
Unit Frob. & $W_1$ & Fixed E20 & Canon. & 0.098 / 0.2649 & 0.436 / 0.0032 & 0.031 / 0.4160 \\
Frob. & Both & Opt. & Raw & 0.155 / 0.2141 & -0.198 / 0.7973 & 0.196 / 0.1373 \\
Frob. & $W_0$ & Opt. & Raw & -0.026 / 0.5700 & 0.153 / 0.2318 & -0.250 / 0.9357 \\
Frob. & $W_1$ & Opt. & Raw & 0.056 / 0.3403 & 0.111 / 0.2622 & 0.418 / 0.0031 \\
Frob. & Balanced & Opt. & Raw & 0.051 / 0.3510 & 0.086 / 0.3139 & 0.420 / 0.0029 \\
Cosine & Both & Opt. & Raw & 0.057 / 0.3699 & 0.178 / 0.1794 & -0.073 / 0.6591 \\
Cosine & $W_0$ & Opt. & Raw & -0.073 / 0.6448 & 0.310 / 0.0194 & -0.243 / 0.9546 \\
Cosine & $W_1$ & Opt. & Raw & 0.109 / 0.2252 & 0.121 / 0.2379 & 0.406 / 0.0042 \\
Unit Frob. & Both & Opt. & Raw & 0.057 / 0.3699 & 0.178 / 0.1794 & -0.073 / 0.6591 \\
Unit Frob. & $W_0$ & Opt. & Raw & -0.073 / 0.6448 & 0.310 / 0.0194 & -0.243 / 0.9546 \\
Unit Frob. & $W_1$ & Opt. & Raw & 0.109 / 0.2252 & 0.121 / 0.2379 & 0.406 / 0.0042 \\
$\alpha_2$ & $W_0$ & $\alpha$ profile & Canon. & 0.079 / 0.3109 & -0.097 / 0.7250 & 0.075 / 0.3171 \\
$\alpha_2$ & $W_1$ & $\alpha$ profile & Canon. & -0.004 / 0.4979 & -0.174 / 0.8102 & -0.287 / 0.9490 \\
$\alpha_2$ & Balanced & $\alpha$ profile & Canon. & 0.066 / 0.3413 & 0.027 / 0.4372 & 0.086 / 0.2840 \\
$\alpha_2$ & Size-wtd. & $\alpha$ profile & Canon. & 0.066 / 0.3417 & -0.097 / 0.7270 & 0.064 / 0.3439 \\
$\alpha_2$ & $W_0$ & $\alpha$ profile & Raw & -0.161 / 0.8094 & -0.147 / 0.8120 & -0.028 / 0.5796 \\
$\alpha_2$ & $W_1$ & $\alpha$ profile & Raw & 0.131 / 0.2196 & -0.280 / 0.9525 & -0.109 / 0.7667 \\
$\alpha_2$ & Balanced & $\alpha$ profile & Raw & 0.055 / 0.3762 & -0.186 / 0.8369 & -0.214 / 0.9079 \\
$\alpha_2$ & Size-wtd. & $\alpha$ profile & Raw & -0.154 / 0.7993 & -0.156 / 0.8223 & -0.032 / 0.5877 \\
SND & $W_0$ & SND profile & Raw & 0.074 / 0.3881 & 0.437 / 0.0148 & 0.141 / 0.1747 \\
SND & $W_1$ & SND profile & Raw & 0.085 / 0.2964 & -0.173 / 0.8693 & -0.028 / 0.5697 \\
SND & Balanced & SND profile & Raw & 0.107 / 0.2485 & -0.087 / 0.7227 & -0.005 / 0.5126 \\
SND & Size-wtd. & SND profile & Raw & 0.084 / 0.3738 & 0.462 / 0.0071 & 0.135 / 0.1799 \\
\end{longtable}\endgroup

The optimal permutation can change over training. Fixing each pair's E0 assignment gives correlations of 0.2986, 0.3585, and $-0.0294$; fixing its E20 assignment gives 0.1559, 0.1663, and 0.3202. The E20-fixed analysis uses endpoint information and is a sensitivity comparison. At the level of individual neuron correspondences, more than 90\% of assignments change on average between E0 and E20. Pairwise optimal matches also fail to define a globally cycle-consistent labeling of every network. Neither observation invalidates pairwise distances, but both explain why assignment conventions matter.

The initial distance distribution is narrow: its coefficient of variation is only 0.58--0.75\%. Precision checks nevertheless preserve all initial and final distance ranks between float32 and float64 calculations. Thus, the narrow spread is a property of the sampled initialization ensemble rather than a demonstrated floating-point artifact.

\begingroup\footnotesize\setlength{\tabcolsep}{4pt}
\renewcommand{\arraystretch}{1.12}
\begin{longtable}{@{}lrrrrr@{}}
\caption{Initial distance concentration, alignment change, and numerical stability. Cycle consistency is the fraction of neuron correspondences agreeing around a network triplet; no complete triplet was exactly consistent (120 triplets per dataset and checkpoint). Rank stability is the minimum float32--float64 Spearman correlation over E0 and E20.}\label{tab:s-rsa-diagnostics}\\
\toprule
Dataset & Initial CV (\%) & Changed (\%) & Cycle E0 (\%) & Cycle E20 (\%) & Rank stability \\
\midrule\endfirsthead
\multicolumn{6}{l}{\textit{Table \thetable\ continued}}\\
\toprule
Dataset & Initial CV (\%) & Changed (\%) & Cycle E0 (\%) & Cycle E20 (\%) & Rank stability \\
\midrule\endhead
\midrule\multicolumn{6}{r}{\textit{Continued on next page}}\\\endfoot
\bottomrule\endlastfoot
MNIST & 0.700 & 92.05 & 14.70 & 26.93 & 1.0000 \\
Fashion & 0.751 & 93.51 & 15.77 & 28.98 & 1.0000 \\
KMNIST & 0.584 & 91.01 & 13.39 & 26.54 & 1.0000 \\
\end{longtable}\endgroup

Norm controls ask whether pairwise distance is associated with the sum or absolute difference of network norms, and whether initial--final RSA persists after conditioning on both. These conditional comparisons are descriptive: fixing covariates during an identity permutation does not create an exact conditional randomization test for arbitrary covariate dependence. Initial--final correlations of individual network norms are reported separately. Exploratory nearest-neighbor retention likewise shows no consistent preservation across datasets.

\begingroup\footnotesize\setlength{\tabcolsep}{4pt}
\renewcommand{\arraystretch}{1.12}
\begin{longtable}{@{}lllllrrr@{}}
\caption{Norm sensitivity. Entries are $\rho$ / unadjusted $p$. Distance rows relate pairwise norms to distances; partial RSA conditions initial--final distance association on both norm predictors. These are descriptive sensitivity tests, not exact conditional tests under arbitrary covariate dependence.}\label{tab:s-rsa-norms}\\
\toprule
Analysis & E & Gauge & Layer & Norm & MNIST & Fashion & KMNIST \\
\midrule\endfirsthead
\multicolumn{8}{l}{\textit{Table \thetable\ continued}}\\
\toprule
Analysis & E & Gauge & Layer & Norm & MNIST & Fashion & KMNIST \\
\midrule\endhead
\midrule\multicolumn{8}{r}{\textit{Continued on next page}}\\\endfoot
\bottomrule\endlastfoot
Distance & 0 & Raw & $W_0$ & Sum & 0.115 / 0.2431 & 0.671 / 0.0004 & 0.371 / 0.0158 \\
Distance & 0 & Raw & $W_0$ & Difference & 0.143 / 0.1912 & -0.383 / 0.9701 & -0.110 / 0.7520 \\
Distance & 20 & Raw & $W_0$ & Sum & 0.179 / 0.2617 & -0.158 / 0.7208 & -0.164 / 0.7344 \\
Distance & 20 & Raw & $W_0$ & Difference & -0.021 / 0.5233 & 0.210 / 0.1464 & -0.157 / 0.7909 \\
Partial RSA & 20 & Raw & $W_0$ & Both & -0.045 / 0.6128 & 0.387 / 0.0033 & -0.192 / 0.9009 \\
Distance & 0 & Raw & $W_1$ & Sum & 0.132 / 0.1182 & 0.189 / 0.1961 & 0.333 / 0.0389 \\
Distance & 0 & Raw & $W_1$ & Difference & 0.315 / 0.0189 & -0.275 / 0.9483 & 0.066 / 0.3698 \\
Distance & 20 & Raw & $W_1$ & Sum & -0.090 / 0.6532 & -0.002 / 0.5128 & 0.029 / 0.4488 \\
Distance & 20 & Raw & $W_1$ & Difference & 0.178 / 0.1563 & 0.108 / 0.2594 & -0.035 / 0.5973 \\
Partial RSA & 20 & Raw & $W_1$ & Both & 0.001 / 0.4869 & 0.148 / 0.1999 & 0.432 / 0.0029 \\
Distance & 0 & Raw & Both & Sum & 0.541 / 0.0011 & 0.731 / 0.0018 & 0.430 / 0.0257 \\
Distance & 0 & Raw & Both & Difference & 0.178 / 0.1367 & 0.031 / 0.4149 & -0.236 / 0.9023 \\
Distance & 20 & Raw & Both & Sum & 0.155 / 0.2809 & -0.363 / 0.9226 & 0.265 / 0.1315 \\
Distance & 20 & Raw & Both & Difference & -0.130 / 0.7326 & 0.094 / 0.2469 & -0.001 / 0.5186 \\
Partial RSA & 20 & Raw & Both & Both & 0.048 / 0.3840 & 0.120 / 0.2758 & 0.103 / 0.2701 \\
Distance & 0 & Canon. & $W_0$ & Sum & 0.904 / 0.0001 & 0.910 / 0.0001 & 0.857 / 0.0003 \\
Distance & 0 & Canon. & $W_0$ & Difference & 0.493 / 0.0079 & -0.125 / 0.7793 & 0.111 / 0.3370 \\
Distance & 20 & Canon. & $W_0$ & Sum & -0.027 / 0.5439 & -0.401 / 0.9424 & 0.236 / 0.1095 \\
Distance & 20 & Canon. & $W_0$ & Difference & 0.017 / 0.4903 & 0.026 / 0.4053 & -0.055 / 0.6089 \\
Partial RSA & 20 & Canon. & $W_0$ & Both & 0.257 / 0.0447 & -0.120 / 0.7791 & 0.087 / 0.2860 \\
Distance & 0 & Canon. & $W_1$ & Sum & 0.455 / 0.0017 & 0.017 / 0.4562 & 0.223 / 0.1021 \\
Distance & 0 & Canon. & $W_1$ & Difference & 0.310 / 0.0299 & -0.073 / 0.6725 & -0.011 / 0.5347 \\
Distance & 20 & Canon. & $W_1$ & Sum & 0.084 / 0.3305 & -0.049 / 0.5933 & -0.012 / 0.5378 \\
Distance & 20 & Canon. & $W_1$ & Difference & 0.189 / 0.1136 & -0.083 / 0.6919 & -0.127 / 0.7786 \\
Partial RSA & 20 & Canon. & $W_1$ & Both & -0.002 / 0.4989 & 0.479 / 0.0009 & -0.009 / 0.5288 \\
Distance & 0 & Canon. & Both & Sum & 0.791 / 0.0002 & 0.602 / 0.0022 & 0.709 / 0.0005 \\
Distance & 0 & Canon. & Both & Difference & 0.473 / 0.0073 & -0.166 / 0.8567 & 0.095 / 0.3503 \\
Distance & 20 & Canon. & Both & Sum & -0.006 / 0.5016 & -0.289 / 0.9121 & 0.160 / 0.1876 \\
Distance & 20 & Canon. & Both & Difference & 0.088 / 0.3298 & -0.033 / 0.5671 & -0.113 / 0.7497 \\
Partial RSA & 20 & Canon. & Both & Both & 0.184 / 0.1010 & -0.017 / 0.5401 & 0.211 / 0.0941 \\
\end{longtable}\endgroup
\begingroup\footnotesize\setlength{\tabcolsep}{4pt}
\renewcommand{\arraystretch}{1.12}
\begin{longtable}{@{}llrrr@{}}
\caption{Initial--final correlation of individual network norms. Entries are Spearman $\rho$ / unadjusted one-sided permutation $p$. These exploratory controls concern norm persistence, rather than origin identification from full weights.}\label{tab:s-norm-persistence}\\
\toprule
Gauge & Layer & MNIST & Fashion & KMNIST \\
\midrule\endfirsthead
\multicolumn{5}{l}{\textit{Table \thetable\ continued}}\\
\toprule
Gauge & Layer & MNIST & Fashion & KMNIST \\
\midrule\endhead
\midrule\multicolumn{5}{r}{\textit{Continued on next page}}\\\endfoot
\bottomrule\endlastfoot
Raw & $W_0$ & 0.261 / 0.2294 & -0.188 / 0.7118 & -0.442 / 0.9074 \\
Raw & $W_1$ & 0.418 / 0.1180 & 0.115 / 0.3750 & -0.527 / 0.9428 \\
Raw & Both & 0.321 / 0.1811 & 0.152 / 0.3470 & 0.333 / 0.1753 \\
Canon. & $W_0$ & 0.382 / 0.1407 & -0.006 / 0.5121 & -0.418 / 0.8930 \\
Canon. & $W_1$ & 0.382 / 0.1407 & -0.006 / 0.5121 & -0.418 / 0.8930 \\
Canon. & Both & 0.382 / 0.1407 & -0.006 / 0.5121 & -0.418 / 0.8930 \\
\end{longtable}\endgroup
\begingroup\footnotesize\setlength{\tabcolsep}{4pt}
\renewcommand{\arraystretch}{1.12}
\begin{longtable}{@{}lrrrrr@{}}
\caption{Exploratory nearest-neighbor retention. Chance retention is $k/9$. Intervals are untruncated network-jackknife intervals and may extend outside the admissible range.}\label{tab:s-neighbors}\\
\toprule
Dataset & $k$ & Retention & Chance & 95\% interval & $p$ \\
\midrule\endfirsthead
\multicolumn{6}{l}{\textit{Table \thetable\ continued}}\\
\toprule
Dataset & $k$ & Retention & Chance & 95\% interval & $p$ \\
\midrule\endhead
\midrule\multicolumn{6}{r}{\textit{Continued on next page}}\\\endfoot
\bottomrule\endlastfoot
MNIST & 1 & 0.300 & 0.111 & [-0.366, 0.966] & 0.1248 \\
MNIST & 3 & 0.367 & 0.333 & [0.056, 0.678] & 0.4163 \\
Fashion & 1 & 0.200 & 0.111 & [-0.192, 0.592] & 0.3021 \\
Fashion & 3 & 0.233 & 0.333 & [-0.132, 0.598] & 0.9105 \\
KMNIST & 1 & 0.000 & 0.111 & [-0.320, 0.320] & 1.0000 \\
KMNIST & 3 & 0.433 & 0.333 & [0.059, 0.808] & 0.1595 \\
\end{longtable}\endgroup

\section{Identification of known initializations (02)}
\label{supp:retrieval}

Every network identifies its own initialization among the ten candidates at all 21 checkpoints. Mean rank and reciprocal rank are both one, and top-1 identification is 100\%, for every dataset and checkpoint. E0 is a self-comparison sanity check; the primary inference concerns E20. The 630 query/checkpoint outcomes reuse 30 networks and do not constitute 630 independent initializations. Retrieval establishes identification among known candidates, not reconstruction of an unknown weight vector.

\begingroup\footnotesize\setlength{\tabcolsep}{4pt}
\renewcommand{\arraystretch}{1.12}
\begin{longtable}{@{}lrrrr@{}}
\caption{Primary E20 weight retrieval. MRR is mean reciprocal rank. All 630 query/checkpoint comparisons also ranked the correct candidate first.}\label{tab:s-retrieval}\\
\toprule
Dataset & Mean rank & Top-1 (\%) & MRR & Holm $p$ \\
\midrule\endfirsthead
\multicolumn{5}{l}{\textit{Table \thetable\ continued}}\\
\toprule
Dataset & Mean rank & Top-1 (\%) & MRR & Holm $p$ \\
\midrule\endhead
\midrule\multicolumn{5}{r}{\textit{Continued on next page}}\\\endfoot
\bottomrule\endlastfoot
MNIST & 1.0 & 100 & 1.0 & 0.0003 \\
Fashion & 1.0 & 100 & 1.0 & 0.0003 \\
KMNIST & 1.0 & 100 & 1.0 & 0.0003 \\
\end{longtable}\endgroup

\section{Complete initialization-control grid (03)}
\label{supp:controls}

The analysis comprises 42 planned control protocols across all 21 checkpoints, supplemented by six post hoc norm-rank protocols. All are exploratory. The complete endpoint grid below includes both gauges, native and optimal coordinates, normalized and unnormalized costs, input and output contributions, activity masks, residual masks, and norm comparisons. Full checkpoint trajectories, intervals, and permutation probabilities are supplied as machine-readable results.

Native coordinates provide a diagnostic of coordinate memory; they do not remove neuron permutations. For canonical comparisons, both layers share the gauge transformation, total-vector normalization, and assignment. Poor output-layer origin retrieval under a joint assignment therefore does not establish loss of all output-layer information. The same output weights remain identifiable in native coordinates.

Low- and high-activity controls select 16 of the 64 hidden units from the current network alone, using the fixed 3,000-image probe. No unit is entirely inactive on the endpoint probe. Small- and large-residual controls select the lower or upper quartile of absolute matched, normalized discrepancies separately for every candidate pair. A small residual under such a comparison is not an unchanged parameter along the actual training history. Historical parameter changes are measured separately below.

\begingroup\footnotesize\setlength{\tabcolsep}{4pt}
\renewcommand{\arraystretch}{1.12}
\begin{longtable}{@{}llllrrr@{}}
\caption{All 48 initialization-control endpoints. Entries are mean correct-candidate rank / top-1 percentage; chance values are 5.5 / 10. The 42 planned comparisons and six post hoc norm-rank comparisons are exploratory. Full intervals and unadjusted permutation probabilities accompany the source table.}\label{tab:s-controls}\\
\toprule
Cost & Layer & Assignment & Gauge & MNIST & Fashion & KMNIST \\
\midrule\endfirsthead
\multicolumn{7}{l}{\textit{Table \thetable\ continued}}\\
\toprule
Cost & Layer & Assignment & Gauge & MNIST & Fashion & KMNIST \\
\midrule\endhead
\midrule\multicolumn{7}{r}{\textit{Continued on next page}}\\\endfoot
\bottomrule\endlastfoot
Frob. & $W_0$ & Native & Raw & 1.00 / 100 & 1.00 / 100 & 1.00 / 100 \\
Frob. & $W_1$ & Native & Raw & 1.00 / 100 & 1.00 / 100 & 1.00 / 100 \\
Frob. & Both & Native & Raw & 1.00 / 100 & 1.00 / 100 & 1.00 / 100 \\
Unit Frob. & $W_0$ & Native & Raw & 1.00 / 100 & 1.00 / 100 & 1.00 / 100 \\
Unit Frob. & $W_1$ & Native & Raw & 1.00 / 100 & 1.00 / 100 & 1.00 / 100 \\
Unit Frob. & Both & Native & Raw & 1.00 / 100 & 1.00 / 100 & 1.00 / 100 \\
Frob. & $W_0$ & Opt. & Raw & 1.00 / 100 & 1.00 / 100 & 1.00 / 100 \\
Frob. & $W_1$ & Opt. & Raw & 1.80 / 80 & 3.00 / 70 & 1.20 / 90 \\
Frob. & Both & Opt. & Raw & 1.00 / 100 & 1.00 / 100 & 1.00 / 100 \\
Unit Frob. & $W_0$ & Opt. & Raw & 1.00 / 100 & 1.00 / 100 & 1.00 / 100 \\
Unit Frob. & $W_1$ & Opt. & Raw & 1.90 / 70 & 2.90 / 60 & 1.10 / 90 \\
Unit Frob. & Both & Opt. & Raw & 1.00 / 100 & 1.00 / 100 & 1.00 / 100 \\
Norm & $W_0$ & None & Raw & 5.50 / 10 & 5.50 / 10 & 5.50 / 10 \\
Norm & $W_1$ & None & Raw & 5.50 / 10 & 5.50 / 10 & 5.50 / 10 \\
Norm & Both & None & Raw & 5.50 / 10 & 5.50 / 10 & 5.50 / 10 \\
Frob. & $W_0$ & Native & Canon. & 1.00 / 100 & 1.00 / 100 & 1.00 / 100 \\
Frob. & $W_1$ & Native & Canon. & 1.00 / 100 & 1.00 / 100 & 1.00 / 100 \\
Frob. & Both & Native & Canon. & 1.00 / 100 & 1.00 / 100 & 1.00 / 100 \\
Unit Frob. & $W_0$ & Native & Canon. & 1.00 / 100 & 1.00 / 100 & 1.00 / 100 \\
Unit Frob. & $W_1$ & Native & Canon. & 1.00 / 100 & 1.00 / 100 & 1.00 / 100 \\
Unit Frob. & Both & Native & Canon. & 1.00 / 100 & 1.00 / 100 & 1.00 / 100 \\
Frob. & $W_0$ & Opt. & Canon. & 1.00 / 100 & 1.00 / 100 & 1.00 / 100 \\
Frob. & $W_1$ & Opt. & Canon. & 8.10 / 0 & 8.80 / 0 & 6.00 / 20 \\
Frob. & Both & Opt. & Canon. & 1.00 / 100 & 1.00 / 100 & 1.00 / 100 \\
Unit Frob. & $W_0$ & Opt. & Canon. & 1.00 / 100 & 1.00 / 100 & 1.00 / 100 \\
Unit Frob. & $W_1$ & Opt. & Canon. & 7.80 / 0 & 8.70 / 0 & 6.00 / 20 \\
Unit Frob. & Both & Opt. & Canon. & 1.00 / 100 & 1.00 / 100 & 1.00 / 100 \\
Norm & $W_0$ & None & Canon. & 5.50 / 10 & 5.50 / 10 & 5.50 / 10 \\
Norm & $W_1$ & None & Canon. & 5.50 / 10 & 5.50 / 10 & 5.50 / 10 \\
Norm & Both & None & Canon. & 5.50 / 10 & 5.50 / 10 & 5.50 / 10 \\
Small resid. & $W_0$ & Opt. & Canon. & 1.00 / 100 & 1.00 / 100 & 1.00 / 100 \\
Small resid. & $W_1$ & Opt. & Canon. & 4.80 / 20 & 5.00 / 10 & 7.00 / 0 \\
Small resid. & Both & Opt. & Canon. & 1.00 / 100 & 1.10 / 90 & 1.10 / 90 \\
Large resid. & $W_0$ & Opt. & Canon. & 1.00 / 100 & 1.00 / 100 & 1.00 / 100 \\
Large resid. & $W_1$ & Opt. & Canon. & 8.30 / 0 & 9.30 / 0 & 6.70 / 0 \\
Large resid. & Both & Opt. & Canon. & 1.00 / 100 & 1.00 / 100 & 1.00 / 100 \\
Low activity & $W_0$ & Masked & Canon. & 1.00 / 100 & 1.00 / 100 & 1.10 / 90 \\
Low activity & $W_1$ & Masked & Canon. & 5.30 / 20 & 6.80 / 20 & 4.70 / 30 \\
Low activity & Both & Masked & Canon. & 1.00 / 100 & 1.10 / 90 & 1.00 / 100 \\
High activity & $W_0$ & Masked & Canon. & 1.00 / 100 & 1.00 / 100 & 1.00 / 100 \\
High activity & $W_1$ & Masked & Canon. & 5.10 / 30 & 6.40 / 0 & 5.30 / 30 \\
High activity & Both & Masked & Canon. & 1.00 / 100 & 1.40 / 90 & 1.00 / 100 \\
Norm rank & $W_0$ & None & Raw & 5.15 / 10 & 5.75 / 10 & 6.60 / 10 \\
Norm rank & $W_1$ & None & Raw & 3.80 / 40 & 5.60 / 10 & 7.10 / 0 \\
Norm rank & Both & None & Raw & 5.10 / 10 & 5.30 / 20 & 4.35 / 30 \\
Norm rank & $W_0$ & None & Canon. & 4.70 / 10 & 6.20 / 10 & 6.65 / 10 \\
Norm rank & $W_1$ & None & Canon. & 4.70 / 10 & 6.20 / 10 & 6.65 / 10 \\
Norm rank & Both & None & Canon. & 4.70 / 10 & 6.20 / 10 & 6.65 / 10 \\
\end{longtable}\endgroup

Norm-rank controls compare a network's position in the initial and final norm ensembles, addressing the tendency of absolute norm growth to rank every candidate in a similar way. They do not reproduce the full-weight signal. Some individual exploratory norm tests are positive, so the conclusion is that global scale does not explain the robust whole-weight identification, rather than that norms contain no information in any condition.

\begingroup\footnotesize\setlength{\tabcolsep}{4pt}
\renewcommand{\arraystretch}{1.12}
\begin{longtable}{@{}llrrr@{}}
\caption{Actual E0--E20 changes in native raw weights, averaged over ten networks per dataset. These historical changes are distinct from candidate-dependent retrieval residuals.}\label{tab:s-changes}\\
\toprule
Dataset & Layer & Relative displacement & Exactly unchanged (\%) & Change $\leq10^{-6}$ (\%) \\
\midrule\endfirsthead
\multicolumn{5}{l}{\textit{Table \thetable\ continued}}\\
\toprule
Dataset & Layer & Relative displacement & Exactly unchanged (\%) & Change $\leq10^{-6}$ (\%) \\
\midrule\endhead
\midrule\multicolumn{5}{r}{\textit{Continued on next page}}\\\endfoot
\bottomrule\endlastfoot
Fashion & $W_0$ & 4.288 & 0.000 & 0.000 \\
Fashion & $W_1$ & 3.896 & 0.000 & 0.000 \\
KMNIST & $W_0$ & 5.053 & 0.000 & 0.001 \\
KMNIST & $W_1$ & 4.926 & 0.000 & 0.000 \\
MNIST & $W_0$ & 3.867 & 0.000 & 0.002 \\
MNIST & $W_1$ & 4.495 & 0.000 & 0.000 \\
\end{longtable}\endgroup

\section{Cross-time representation retrieval (04)}
\label{supp:representations}

Raw hidden CKA, standardized hidden CKA, and raw-logit CKA evaluated origin identification on the same fixed probe. The aggregate final mean ranks are 5.40, 5.53, and 5.60, respectively, against chance rank 5.5. Feature standardization tests sensitivity to positive featurewise scaling; it does not equate all representational comparisons. The complete endpoint results show no consistent correct-origin advantage. This failed cross-time retrieval does not establish that every aspect of initial representational information was erased. Experiment 05 tests the distinct proposition that a shared initialization can shape newly learned final representations.

\begingroup\footnotesize\setlength{\tabcolsep}{4pt}
\renewcommand{\arraystretch}{1.12}
\begin{longtable}{@{}lllrrrr@{}}
\caption{All E20 representation-retrieval conditions. Hidden raw CKA is primary; feature-standardized hidden CKA and logits are secondary. Mean-rank intervals and one-sided probabilities retain the source analysis conventions; displayed probabilities are unadjusted.}\label{tab:s-representations}\\
\toprule
Dataset & Features & Scale & Rank & 95\% interval & Top-1 (\%) & $p$ \\
\midrule\endfirsthead
\multicolumn{7}{l}{\textit{Table \thetable\ continued}}\\
\toprule
Dataset & Features & Scale & Rank & 95\% interval & Top-1 (\%) & $p$ \\
\midrule\endhead
\midrule\multicolumn{7}{r}{\textit{Continued on next page}}\\\endfoot
\bottomrule\endlastfoot
MNIST & hidden & raw & 5.500 & [4.570, 6.430] & 0.0 & 0.5417 \\
MNIST & hidden & standardized & 5.300 & [4.418, 6.182] & 10.0 & 0.3568 \\
MNIST & logits & raw & 5.600 & [5.159, 6.041] & 10.0 & 0.6456 \\
Fashion & hidden & raw & 4.800 & [3.519, 6.081] & 10.0 & 0.0754 \\
Fashion & hidden & standardized & 5.200 & [4.840, 5.560] & 10.0 & 0.0950 \\
Fashion & logits & raw & 5.400 & [4.608, 6.192] & 20.0 & 0.4208 \\
KMNIST & hidden & raw & 5.900 & [4.820, 6.980] & 10.0 & 0.8453 \\
KMNIST & hidden & standardized & 6.100 & [5.243, 6.957] & 10.0 & 0.9997 \\
KMNIST & logits & raw & 5.800 & [5.050, 6.550] & 10.0 & 0.9008 \\
Aggregate & hidden & raw & 5.400 & [4.761, 6.039] & 6.7 & 0.3669 \\
Aggregate & hidden & standardized & 5.533 & [5.106, 5.960] & 10.0 & 0.6084 \\
Aggregate & logits & raw & 5.600 & [5.208, 5.992] & 13.3 & 0.7363 \\
\end{longtable}\endgroup

\section{Crossed initialization and batch schedule (05)}
\label{supp:crossed}

The 150 newly trained networks cross ten original initializations with five independently specified batch schedules in each of three datasets. They add schedules, not 150 independent initializations. Initialization retrieval averages 20 directed schedule-pair comparisons within each initialization before inference; schedule retrieval analogously averages 90 directed initialization-pair comparisons. Both weight and hidden-CKA results are retained in the main figures. All 150 endpoints also identify their original E0 weights in the initial--final retrieval control.

For retrieval of the batch schedule from final weights, top-1 accuracy is the percentage of queries for which the correct schedule ranks first among five candidates (chance, 20\%). Across all ten initializations, the correct schedule ranked first in 110 of 450 comparisons on MNIST (24.44\%), 149 of 450 on Fashion-MNIST (33.11\%), and 137 of 450 on KMNIST (30.44\%). The deletion ranges in Table~\ref{tab:s-factor-deletion} describe sensitivity to omitting one initialization.

The geometric contrasts reinforce the retrieval results: sharing either factor reduces average distance relative to pairs differing in both. There are 100 unordered same-initialization pairs, 225 same-schedule pairs, and 900 pairs differing in both factors per dataset. These pairs overlap and are not independent replicates. Inference uses the source analysis's factor-block randomization and deletion procedures.

\begingroup\footnotesize\setlength{\tabcolsep}{4pt}
\renewcommand{\arraystretch}{1.12}
\begin{longtable}{@{}lllrrr@{}}
\caption{Geometric contrasts: distance with both factors different minus distance with the named factor shared. Positive values indicate closer endpoints when that factor is shared. Factor-block intervals and unadjusted secondary permutation tests are retained.}\label{tab:s-contrasts}\\
\toprule
Dataset & Space & Shared factor & Contrast & 95\% interval & $p$ \\
\midrule\endfirsthead
\multicolumn{6}{l}{\textit{Table \thetable\ continued}}\\
\toprule
Dataset & Space & Shared factor & Contrast & 95\% interval & $p$ \\
\midrule\endhead
\midrule\multicolumn{6}{r}{\textit{Continued on next page}}\\\endfoot
\bottomrule\endlastfoot
MNIST & Weights & Init. & 0.1669 & [0.1560, 0.1777] & 0.0001 \\
MNIST & Weights & Schedule & 0.0019 & [0.0008, 0.0029] & 0.0031 \\
MNIST & CKA & Init. & 0.0446 & [0.0393, 0.0498] & 0.0001 \\
MNIST & CKA & Schedule & 0.0027 & [0.0013, 0.0042] & 0.0001 \\
Fashion & Weights & Init. & 0.0688 & [0.0584, 0.0793] & 0.0001 \\
Fashion & Weights & Schedule & 0.0074 & [0.0045, 0.0102] & 0.0001 \\
Fashion & CKA & Init. & 0.0209 & [0.0165, 0.0253] & 0.0001 \\
Fashion & CKA & Schedule & 0.0078 & [0.0054, 0.0101] & 0.0001 \\
KMNIST & Weights & Init. & 0.1431 & [0.1318, 0.1544] & 0.0001 \\
KMNIST & Weights & Schedule & 0.0033 & [0.0016, 0.0050] & 0.0001 \\
KMNIST & CKA & Init. & 0.0496 & [0.0448, 0.0544] & 0.0001 \\
KMNIST & CKA & Schedule & 0.0041 & [0.0023, 0.0059] & 0.0001 \\
\end{longtable}\endgroup
\begingroup\footnotesize\setlength{\tabcolsep}{4pt}
\renewcommand{\arraystretch}{1.12}
\begin{longtable}{@{}lllrr@{}}
\caption{Deletion of levels of the other factor. Ranges summarize the recorded leave-one-level estimates; they are sensitivity ranges, not confidence intervals. Initialization retrieval deletes a schedule; schedule retrieval deletes an initialization.}\label{tab:s-factor-deletion}\\
\toprule
Dataset & Identify & Space & Rank range & Top-1 range (\%) \\
\midrule\endfirsthead
\multicolumn{5}{l}{\textit{Table \thetable\ continued}}\\
\toprule
Dataset & Identify & Space & Rank range & Top-1 range (\%) \\
\midrule\endhead
\midrule\multicolumn{5}{r}{\textit{Continued on next page}}\\\endfoot
\bottomrule\endlastfoot
Fashion & Init. & Weights & [1.025, 1.325] & [85.8, 98.3] \\
Fashion & Init. & CKA & [2.650, 2.925] & [41.7, 50.8] \\
Fashion & Schedule & Weights & [2.439, 2.589] & [30.8, 35.6] \\
Fashion & Schedule & CKA & [2.400, 2.506] & [32.5, 36.4] \\
KMNIST & Init. & Weights & [1.000, 1.000] & [100.0, 100.0] \\
KMNIST & Init. & CKA & [1.175, 1.358] & [85.0, 90.0] \\
KMNIST & Schedule & Weights & [2.544, 2.683] & [27.5, 33.3] \\
KMNIST & Schedule & CKA & [2.603, 2.714] & [26.4, 30.0] \\
MNIST & Init. & Weights & [1.000, 1.000] & [100.0, 100.0] \\
MNIST & Init. & CKA & [1.042, 1.083] & [96.7, 97.5] \\
MNIST & Schedule & Weights & [2.775, 2.881] & [22.5, 25.8] \\
MNIST & Schedule & CKA & [2.669, 2.761] & [25.8, 29.4] \\
\end{longtable}\endgroup

Structural identity and performance variation answer different questions. Mean final accuracies are 97.27\%, 86.38\%, and 86.05\% for MNIST, Fashion-MNIST, and KMNIST. The descriptive sums-of-squares decomposition assigns 18.1\%, 11.1\%, and 19.9\% of observed accuracy variation to the additive initialization term, versus 16.6\%, 43.8\%, and 26.6\% to the schedule term. These fractions have different degrees of freedom (9, 4, and 36 for initialization, schedule, and interaction/rest). With one observation per cell, interaction and additional residual error cannot be separated, and these fractions are not unbiased estimates of random variance components. They establish why strong origin identification should not be equated with initialization dominating predictive performance.

\begingroup\footnotesize\setlength{\tabcolsep}{4pt}
\renewcommand{\arraystretch}{1.12}
\begin{longtable}{@{}llrrrr@{}}
\caption{Descriptive crossed-design performance decomposition at E20. Fractions partition the observed sum of squares, not unbiased random-effect variance components. One observation per cell prevents separation of interaction and residual error; no ANOVA tests are asserted.}\label{tab:s-performance}\\
\toprule
Dataset & Measure & Mean & Init. fraction & Schedule fraction & Interaction/rest \\
\midrule\endfirsthead
\multicolumn{6}{l}{\textit{Table \thetable\ continued}}\\
\toprule
Dataset & Measure & Mean & Init. fraction & Schedule fraction & Interaction/rest \\
\midrule\endhead
\midrule\multicolumn{6}{r}{\textit{Continued on next page}}\\\endfoot
\bottomrule\endlastfoot
MNIST & accuracy & 0.9727 & 0.181 & 0.166 & 0.654 \\
MNIST & loss & 0.1150 & 0.195 & 0.134 & 0.671 \\
Fashion & accuracy & 0.8638 & 0.111 & 0.438 & 0.451 \\
Fashion & loss & 0.4041 & 0.084 & 0.302 & 0.614 \\
KMNIST & accuracy & 0.8605 & 0.199 & 0.266 & 0.535 \\
KMNIST & loss & 0.7921 & 0.249 & 0.250 & 0.502 \\
\end{longtable}\endgroup

\section{Completed perturbation study and dynamical controls (06)}
\label{supp:perturbation}

\subsection{Scope and operational definitions}

The MNIST pilot contains 300 perturbed continuations, 20 same-order unperturbed continuations, and 60 changed-order controls. Ten independent source networks at E20 each contribute five directions per nonzero strength and momentum condition. The extension contains all 2,160 planned perturbations and all 180 matched baselines, with zero numerical failures. It crosses three datasets, E2/E10/E20, three layer choices, four strengths, two momentum protocols, and ten sources, with one direction per source and cell. Each continuation uses ten further epochs. Its 90 independent reference continuations reach matching horizons; source states and reference initializations are reused across intervention times.

The primary ratio is the canonical, jointly permutation-matched distance at the end divided by the distance immediately after intervention. A ratio below one denotes contraction relative to that immediate displacement. Retained and reset momentum are applied symmetrically to both branches. The pilot averages directions within a source before resampling ten sources; the extension resamples its ten sources directly. All reported ratio intervals are pointwise, descriptive 95\% bootstrap intervals rather than simultaneous confidence statements about the full grid.

A reference region combines three criteria: median structural distance to reference networks below the 95th percentile of their leave-one-out median distances, accuracy above its calibrated reference threshold, and loss below its reference threshold. The time-matched envelope changes with training time. The fixed region uses the final-horizon references and thresholds at every measurement time. A perturbation-induced return requires that the pre-intervention baseline was inside, the immediately perturbed state was outside, and the final observation was inside again. The structural outside-to-inside diagnostic requires an immediately perturbed state outside the structural envelope and a final state inside it; it does not impose the additional pre-intervention membership condition. These definitions distinguish endpoint return indicators from transient intermediate entries and from ordinary progress toward a later training stage.

For the pilot, the final structural threshold is 20.693572, the alternative 90th-percentile threshold is 20.682505, the accuracy threshold is 97.467\%, and the loss threshold is 0.146274. The local-nearness criterion uses radius 2.060010 around the paired baseline together with the functional criteria. Reference thresholds are empirical estimates from ten networks; bootstrap intervals over intervention sources do not incorporate uncertainty in these reference boundaries.

\subsection{Complete expanded endpoint grid}

All 216 expanded cells have mean ratios above one and pointwise lower bounds above one. The tables below retain every cell and both momentum protocols, alongside final accuracy and fixed combined-region membership. This result holds for individual-layer and joint perturbations at all three intervention times. A larger relative ratio for a smaller perturbation does not imply a larger absolute endpoint distance: the denominator changes with intervention size. Absolute distances, complete recovery curves, and their source-level intervals accompany the endpoint tables in the numerical archive.

\clearpage
\begingroup\footnotesize\setlength{\tabcolsep}{4pt}
\renewcommand{\arraystretch}{1.12}
\begin{longtable}{@{}llrrrrr@{}}
\caption{Complete extension: MNIST. Each row represents two momentum conditions, each with ten sources and one direction per source. Ratios have pointwise 95\% bootstrap intervals. Accuracy and fixed combined-region membership are percentages, shown retained / reset. E2, E10, and E20 are one, five, and ten epochs before ten further epochs.}\label{tab:s-expanded-mnist}\\
\toprule
E & Layer & $\epsilon$ (\%) & $r$: retained [CI] & $r$: reset [CI] & Accuracy R/S & Region R/S \\
\midrule\endfirsthead
\multicolumn{7}{l}{\textit{Table \thetable\ continued}}\\
\toprule
E & Layer & $\epsilon$ (\%) & $r$: retained [CI] & $r$: reset [CI] & Accuracy R/S & Region R/S \\
\midrule\endhead
\midrule\multicolumn{7}{r}{\textit{Continued on next page}}\\\endfoot
\bottomrule\endlastfoot
2 & $W_0$ & 1 & 60.29 [58.79, 61.58] & 59.89 [57.78, 61.92] & 97.36 / 97.30 & 90 / 90 \\
2 & $W_0$ & 5 & 12.33 [11.95, 12.70] & 12.31 [11.92, 12.67] & 97.21 / 97.28 & 90 / 90 \\
2 & $W_0$ & 10 & 6.29 [6.09, 6.48] & 6.32 [6.13, 6.52] & 97.33 / 97.32 & 80 / 80 \\
2 & $W_0$ & 20 & 3.36 [3.27, 3.45] & 3.39 [3.30, 3.48] & 97.32 / 97.29 & 80 / 80 \\
2 & $W_1$ & 1 & 59.89 [57.72, 61.94] & 58.87 [56.66, 60.78] & 97.27 / 97.26 & 90 / 80 \\
2 & $W_1$ & 5 & 12.36 [11.96, 12.80] & 12.16 [11.87, 12.41] & 97.35 / 97.23 & 90 / 90 \\
2 & $W_1$ & 10 & 6.31 [6.07, 6.57] & 6.24 [6.09, 6.40] & 97.38 / 97.26 & 80 / 80 \\
2 & $W_1$ & 20 & 3.46 [3.33, 3.59] & 3.48 [3.38, 3.58] & 97.25 / 97.21 & 70 / 60 \\
2 & Both & 1 & 42.66 [41.79, 43.62] & 42.12 [40.87, 43.45] & 97.29 / 97.24 & 80 / 90 \\
2 & Both & 5 & 8.74 [8.42, 9.05] & 8.80 [8.48, 9.09] & 97.29 / 97.36 & 80 / 100 \\
2 & Both & 10 & 4.53 [4.40, 4.66] & 4.62 [4.49, 4.74] & 97.30 / 97.34 & 80 / 70 \\
2 & Both & 20 & 2.57 [2.52, 2.62] & 2.60 [2.55, 2.65] & 97.31 / 97.38 & 70 / 70 \\
10 & $W_0$ & 1 & 32.89 [31.79, 34.08] & 32.88 [31.94, 33.78] & 97.42 / 97.35 & 90 / 70 \\
10 & $W_0$ & 5 & 6.72 [6.51, 6.91] & 6.67 [6.49, 6.83] & 97.47 / 97.48 & 70 / 80 \\
10 & $W_0$ & 10 & 3.49 [3.36, 3.61] & 3.48 [3.37, 3.59] & 97.34 / 97.46 & 70 / 90 \\
10 & $W_0$ & 20 & 1.96 [1.91, 2.02] & 1.96 [1.91, 2.01] & 97.42 / 97.42 & 50 / 60 \\
10 & $W_1$ & 1 & 33.31 [32.22, 34.47] & 33.01 [32.10, 33.87] & 97.47 / 97.39 & 80 / 90 \\
10 & $W_1$ & 5 & 6.67 [6.49, 6.87] & 6.75 [6.51, 6.98] & 97.49 / 97.38 & 90 / 90 \\
10 & $W_1$ & 10 & 3.50 [3.40, 3.60] & 3.48 [3.38, 3.58] & 97.42 / 97.36 & 90 / 80 \\
10 & $W_1$ & 20 & 1.86 [1.81, 1.91] & 1.85 [1.80, 1.90] & 97.37 / 97.39 & 90 / 90 \\
10 & Both & 1 & 23.22 [22.39, 24.07] & 23.33 [22.45, 24.17] & 97.32 / 97.38 & 70 / 70 \\
10 & Both & 5 & 4.84 [4.67, 5.01] & 4.81 [4.63, 4.98] & 97.38 / 97.42 & 80 / 90 \\
10 & Both & 10 & 2.54 [2.48, 2.61] & 2.54 [2.46, 2.61] & 97.39 / 97.32 & 70 / 60 \\
10 & Both & 20 & 1.47 [1.45, 1.50] & 1.48 [1.45, 1.51] & 97.34 / 97.41 & 50 / 60 \\
20 & $W_0$ & 1 & 22.49 [20.76, 24.18] & 22.10 [20.02, 23.98] & 97.59 / 97.57 & 70 / 70 \\
20 & $W_0$ & 5 & 4.59 [4.26, 4.92] & 4.55 [4.12, 4.91] & 97.59 / 97.57 & 70 / 50 \\
20 & $W_0$ & 10 & 2.46 [2.32, 2.60] & 2.46 [2.29, 2.62] & 97.67 / 97.56 & 90 / 60 \\
20 & $W_0$ & 20 & 1.54 [1.48, 1.59] & 1.54 [1.47, 1.59] & 97.51 / 97.58 & 40 / 40 \\
20 & $W_1$ & 1 & 22.75 [20.93, 24.53] & 22.37 [20.07, 24.29] & 97.66 / 97.61 & 90 / 70 \\
20 & $W_1$ & 5 & 4.61 [4.26, 4.97] & 4.54 [4.15, 4.88] & 97.60 / 97.61 & 80 / 70 \\
20 & $W_1$ & 10 & 2.41 [2.24, 2.60] & 2.39 [2.19, 2.57] & 97.65 / 97.55 & 80 / 60 \\
20 & $W_1$ & 20 & 1.33 [1.27, 1.39] & 1.32 [1.25, 1.38] & 97.61 / 97.63 & 60 / 60 \\
20 & Both & 1 & 16.19 [14.91, 17.45] & 15.66 [14.05, 17.02] & 97.63 / 97.62 & 70 / 70 \\
20 & Both & 5 & 3.34 [3.12, 3.55] & 3.25 [2.97, 3.50] & 97.55 / 97.64 & 60 / 70 \\
20 & Both & 10 & 1.83 [1.73, 1.93] & 1.80 [1.69, 1.89] & 97.61 / 97.61 & 80 / 70 \\
20 & Both & 20 & 1.17 [1.13, 1.21] & 1.16 [1.11, 1.20] & 97.53 / 97.60 & 30 / 50 \\
\end{longtable}\endgroup
\begingroup\footnotesize\setlength{\tabcolsep}{4pt}
\renewcommand{\arraystretch}{1.12}
\begin{longtable}{@{}llrrrrr@{}}
\caption{Complete extension: Fashion. Each row represents two momentum conditions, each with ten sources and one direction per source. Ratios have pointwise 95\% bootstrap intervals. Accuracy and fixed combined-region membership are percentages, shown retained / reset. E2, E10, and E20 are one, five, and ten epochs before ten further epochs.}\label{tab:s-expanded-fashion_mnist}\\
\toprule
E & Layer & $\epsilon$ (\%) & $r$: retained [CI] & $r$: reset [CI] & Accuracy R/S & Region R/S \\
\midrule\endfirsthead
\multicolumn{7}{l}{\textit{Table \thetable\ continued}}\\
\toprule
E & Layer & $\epsilon$ (\%) & $r$: retained [CI] & $r$: reset [CI] & Accuracy R/S & Region R/S \\
\midrule\endhead
\midrule\multicolumn{7}{r}{\textit{Continued on next page}}\\\endfoot
\bottomrule\endlastfoot
2 & $W_0$ & 1 & 68.55 [61.14, 75.60] & 72.31 [63.85, 80.60] & 86.75 / 86.70 & 60 / 70 \\
2 & $W_0$ & 5 & 13.94 [12.33, 15.49] & 14.49 [12.61, 16.28] & 86.74 / 86.65 & 60 / 80 \\
2 & $W_0$ & 10 & 7.27 [6.51, 8.05] & 7.38 [6.47, 8.27] & 86.81 / 86.75 & 50 / 80 \\
2 & $W_0$ & 20 & 3.92 [3.48, 4.31] & 3.99 [3.61, 4.34] & 86.84 / 86.75 & 60 / 60 \\
2 & $W_1$ & 1 & 67.42 [60.19, 74.06] & 70.39 [62.86, 78.57] & 86.84 / 86.67 & 70 / 80 \\
2 & $W_1$ & 5 & 13.90 [12.26, 15.43] & 14.02 [12.59, 15.38] & 86.98 / 86.66 & 70 / 70 \\
2 & $W_1$ & 10 & 6.98 [6.29, 7.63] & 7.47 [6.59, 8.35] & 86.81 / 86.64 & 70 / 60 \\
2 & $W_1$ & 20 & 3.92 [3.50, 4.33] & 4.08 [3.72, 4.42] & 86.66 / 86.65 & 70 / 70 \\
2 & Both & 1 & 50.60 [45.55, 55.91] & 52.23 [45.66, 58.37] & 86.71 / 86.70 & 70 / 70 \\
2 & Both & 5 & 10.05 [9.10, 10.97] & 10.04 [8.71, 11.41] & 86.66 / 86.68 & 70 / 60 \\
2 & Both & 10 & 5.25 [4.63, 5.82] & 5.38 [4.75, 6.01] & 86.65 / 86.58 & 70 / 70 \\
2 & Both & 20 & 2.92 [2.69, 3.14] & 3.05 [2.79, 3.32] & 86.72 / 86.67 & 50 / 60 \\
10 & $W_0$ & 1 & 31.85 [29.76, 33.94] & 31.45 [29.61, 33.12] & 86.74 / 86.86 & 80 / 100 \\
10 & $W_0$ & 5 & 6.51 [6.09, 6.92] & 6.46 [6.09, 6.80] & 86.80 / 86.86 & 100 / 90 \\
10 & $W_0$ & 10 & 3.42 [3.21, 3.61] & 3.34 [3.17, 3.50] & 86.75 / 86.89 & 80 / 90 \\
10 & $W_0$ & 20 & 1.93 [1.84, 2.01] & 1.90 [1.81, 1.97] & 86.75 / 86.81 & 70 / 70 \\
10 & $W_1$ & 1 & 32.33 [30.47, 34.14] & 32.35 [29.98, 34.39] & 86.65 / 86.65 & 90 / 90 \\
10 & $W_1$ & 5 & 6.62 [6.22, 6.99] & 6.69 [6.23, 7.12] & 86.65 / 86.89 & 90 / 100 \\
10 & $W_1$ & 10 & 3.38 [3.20, 3.57] & 3.40 [3.16, 3.61] & 86.61 / 86.81 & 100 / 100 \\
10 & $W_1$ & 20 & 1.86 [1.76, 1.95] & 1.86 [1.78, 1.93] & 86.78 / 86.64 & 80 / 80 \\
10 & Both & 1 & 22.88 [21.26, 24.32] & 22.34 [20.94, 23.66] & 86.81 / 86.82 & 90 / 100 \\
10 & Both & 5 & 4.76 [4.52, 4.99] & 4.64 [4.38, 4.89] & 86.85 / 86.86 & 100 / 90 \\
10 & Both & 10 & 2.47 [2.33, 2.61] & 2.47 [2.34, 2.59] & 86.67 / 86.87 & 80 / 80 \\
10 & Both & 20 & 1.48 [1.43, 1.52] & 1.46 [1.40, 1.51] & 86.83 / 86.83 & 80 / 80 \\
20 & $W_0$ & 1 & 23.02 [21.80, 24.22] & 22.90 [21.87, 23.87] & 86.91 / 86.77 & 70 / 70 \\
20 & $W_0$ & 5 & 4.73 [4.53, 4.92] & 4.73 [4.51, 4.94] & 86.83 / 86.98 & 60 / 70 \\
20 & $W_0$ & 10 & 2.49 [2.37, 2.60] & 2.51 [2.42, 2.59] & 86.99 / 86.94 & 70 / 60 \\
20 & $W_0$ & 20 & 1.50 [1.45, 1.55] & 1.51 [1.47, 1.55] & 86.75 / 86.93 & 50 / 50 \\
20 & $W_1$ & 1 & 23.56 [22.45, 24.62] & 24.10 [22.98, 25.15] & 86.92 / 86.91 & 60 / 50 \\
20 & $W_1$ & 5 & 4.84 [4.57, 5.08] & 4.89 [4.66, 5.10] & 86.88 / 86.92 & 50 / 50 \\
20 & $W_1$ & 10 & 2.53 [2.40, 2.65] & 2.52 [2.41, 2.64] & 86.94 / 87.00 & 50 / 60 \\
20 & $W_1$ & 20 & 1.40 [1.35, 1.44] & 1.41 [1.37, 1.45] & 87.04 / 86.96 & 70 / 60 \\
20 & Both & 1 & 16.39 [15.62, 17.08] & 16.63 [15.83, 17.45] & 86.75 / 86.92 & 40 / 40 \\
20 & Both & 5 & 3.43 [3.24, 3.61] & 3.44 [3.30, 3.57] & 86.99 / 86.87 & 70 / 60 \\
20 & Both & 10 & 1.85 [1.78, 1.91] & 1.87 [1.80, 1.93] & 86.89 / 86.99 & 60 / 80 \\
20 & Both & 20 & 1.18 [1.15, 1.20] & 1.18 [1.16, 1.21] & 86.80 / 86.89 & 60 / 70 \\
\end{longtable}\endgroup
\begingroup\footnotesize\setlength{\tabcolsep}{4pt}
\renewcommand{\arraystretch}{1.12}
\begin{longtable}{@{}llrrrrr@{}}
\caption{Complete extension: KMNIST. Each row represents two momentum conditions, each with ten sources and one direction per source. Ratios have pointwise 95\% bootstrap intervals. Accuracy and fixed combined-region membership are percentages, shown retained / reset. E2, E10, and E20 are one, five, and ten epochs before ten further epochs.}\label{tab:s-expanded-kmnist}\\
\toprule
E & Layer & $\epsilon$ (\%) & $r$: retained [CI] & $r$: reset [CI] & Accuracy R/S & Region R/S \\
\midrule\endfirsthead
\multicolumn{7}{l}{\textit{Table \thetable\ continued}}\\
\toprule
E & Layer & $\epsilon$ (\%) & $r$: retained [CI] & $r$: reset [CI] & Accuracy R/S & Region R/S \\
\midrule\endhead
\midrule\multicolumn{7}{r}{\textit{Continued on next page}}\\\endfoot
\bottomrule\endlastfoot
2 & $W_0$ & 1 & 78.85 [76.77, 80.90] & 79.15 [77.30, 80.83] & 85.97 / 85.98 & 90 / 90 \\
2 & $W_0$ & 5 & 16.18 [15.79, 16.64] & 16.24 [15.92, 16.59] & 85.99 / 86.06 & 60 / 90 \\
2 & $W_0$ & 10 & 8.22 [8.09, 8.35] & 8.37 [8.13, 8.62] & 86.19 / 86.13 & 100 / 90 \\
2 & $W_0$ & 20 & 4.32 [4.25, 4.40] & 4.40 [4.31, 4.50] & 86.33 / 86.04 & 90 / 90 \\
2 & $W_1$ & 1 & 77.99 [76.20, 79.80] & 78.58 [76.56, 80.83] & 86.25 / 86.26 & 100 / 100 \\
2 & $W_1$ & 5 & 15.94 [15.58, 16.37] & 16.28 [15.83, 16.75] & 86.15 / 86.04 & 90 / 100 \\
2 & $W_1$ & 10 & 8.38 [8.16, 8.61] & 8.33 [8.14, 8.56] & 86.09 / 86.11 & 80 / 90 \\
2 & $W_1$ & 20 & 4.46 [4.35, 4.59] & 4.48 [4.41, 4.55] & 86.11 / 86.05 & 80 / 100 \\
2 & Both & 1 & 55.13 [53.63, 56.56] & 56.47 [54.93, 58.05] & 86.02 / 86.16 & 80 / 100 \\
2 & Both & 5 & 11.44 [11.21, 11.66] & 11.50 [11.22, 11.78] & 86.06 / 86.15 & 90 / 100 \\
2 & Both & 10 & 5.93 [5.79, 6.08] & 6.05 [5.92, 6.21] & 86.28 / 86.16 & 100 / 80 \\
2 & Both & 20 & 3.25 [3.18, 3.32] & 3.27 [3.21, 3.35] & 86.06 / 85.91 & 80 / 80 \\
10 & $W_0$ & 1 & 44.68 [43.94, 45.37] & 45.06 [44.04, 46.13] & 86.04 / 86.00 & 90 / 90 \\
10 & $W_0$ & 5 & 9.06 [8.88, 9.24] & 9.11 [8.96, 9.27] & 86.04 / 86.06 & 100 / 90 \\
10 & $W_0$ & 10 & 4.66 [4.60, 4.72] & 4.67 [4.59, 4.75] & 86.08 / 86.28 & 100 / 100 \\
10 & $W_0$ & 20 & 2.49 [2.46, 2.52] & 2.49 [2.45, 2.54] & 86.24 / 86.10 & 90 / 90 \\
10 & $W_1$ & 1 & 44.98 [44.32, 45.61] & 45.35 [44.35, 46.37] & 85.98 / 86.11 & 90 / 90 \\
10 & $W_1$ & 5 & 9.13 [8.95, 9.30] & 9.23 [9.06, 9.39] & 86.26 / 86.36 & 100 / 100 \\
10 & $W_1$ & 10 & 4.63 [4.55, 4.71] & 4.65 [4.56, 4.74] & 86.04 / 86.13 & 100 / 90 \\
10 & $W_1$ & 20 & 2.41 [2.37, 2.46] & 2.42 [2.38, 2.46] & 86.09 / 85.98 & 80 / 80 \\
10 & Both & 1 & 31.63 [31.12, 32.20] & 32.12 [31.43, 32.82] & 86.20 / 85.98 & 100 / 90 \\
10 & Both & 5 & 6.50 [6.39, 6.61] & 6.49 [6.36, 6.63] & 86.40 / 86.31 & 100 / 100 \\
10 & Both & 10 & 3.32 [3.27, 3.37] & 3.33 [3.27, 3.39] & 86.01 / 86.26 & 90 / 100 \\
10 & Both & 20 & 1.81 [1.79, 1.83] & 1.81 [1.79, 1.84] & 85.94 / 86.18 & 60 / 80 \\
20 & $W_0$ & 1 & 36.52 [35.93, 37.15] & 37.03 [36.12, 38.10] & 86.12 / 86.28 & 100 / 100 \\
20 & $W_0$ & 5 & 7.41 [7.23, 7.61] & 7.45 [7.23, 7.67] & 86.06 / 86.30 & 90 / 90 \\
20 & $W_0$ & 10 & 3.80 [3.72, 3.88] & 3.85 [3.77, 3.93] & 86.27 / 86.27 & 100 / 100 \\
20 & $W_0$ & 20 & 2.07 [2.03, 2.11] & 2.08 [2.04, 2.12] & 86.35 / 86.19 & 90 / 90 \\
20 & $W_1$ & 1 & 37.08 [36.14, 38.05] & 37.09 [35.99, 38.16] & 86.20 / 86.29 & 100 / 100 \\
20 & $W_1$ & 5 & 7.44 [7.26, 7.63] & 7.55 [7.28, 7.83] & 86.37 / 86.12 & 100 / 90 \\
20 & $W_1$ & 10 & 3.79 [3.67, 3.90] & 3.82 [3.71, 3.92] & 86.34 / 86.04 & 90 / 90 \\
20 & $W_1$ & 20 & 1.95 [1.89, 2.01] & 1.97 [1.92, 2.03] & 86.25 / 86.15 & 100 / 100 \\
20 & Both & 1 & 25.87 [25.25, 26.53] & 26.10 [25.45, 26.77] & 86.28 / 86.30 & 100 / 100 \\
20 & Both & 5 & 5.29 [5.16, 5.43] & 5.34 [5.16, 5.51] & 86.13 / 86.08 & 100 / 90 \\
20 & Both & 10 & 2.73 [2.68, 2.78] & 2.76 [2.69, 2.83] & 86.08 / 86.25 & 100 / 80 \\
20 & Both & 20 & 1.49 [1.46, 1.52] & 1.51 [1.47, 1.54] & 86.45 / 86.27 & 90 / 70 \\
\end{longtable}\endgroup

All 2,160 immediately perturbed states remain inside the fixed structural envelope. There are no observed structural exit--return sequences caused by the immediate perturbation. The 115 perturbation-induced returns to the combined region concern functional criteria while structural membership was already present. These observations neither establish nor refute attraction after leaving a broader structural solution set.

Aligned endpoint interpolations use eleven equally spaced points and the full test set. The barrier is the maximum loss above the linear interpolation of endpoint losses. Of 2,160 perturbed extension pairs, 2,157 have zero positive sampled barrier and three have small positive barriers, with maximum 0.0036138. The discrete sample does not bound the continuous barrier between sample points. The pilot has no positive sampled barrier in its 300 perturbed pairs. Endpoint connectivity and restoring dynamics remain distinct measurements.

\begingroup\footnotesize\setlength{\tabcolsep}{4pt}
\renewcommand{\arraystretch}{1.12}
\begin{longtable}{@{}llrrrrr@{}}
\caption{Expanded functional and connectivity controls. Accuracy and loss ranges span the 24 cell means per dataset/time, each relative to its paired baseline (accuracy in percentage points). CKA ranges likewise span cell means. Returns count individual perturbed runs re-entering the combined region after intervention-induced exit; none is a structural exit--return. Maximum barriers are across the 240 perturbations per dataset/time.}\label{tab:s-expanded-function}\\
\toprule
Dataset & E & Returns & Accuracy $\Delta$ (pp) & Loss $\Delta$ & CKA range & Max. barrier \\
\midrule\endfirsthead
\multicolumn{7}{l}{\textit{Table \thetable\ continued}}\\
\toprule
Dataset & E & Returns & Accuracy $\Delta$ (pp) & Loss $\Delta$ & CKA range & Max. barrier \\
\midrule\endhead
\midrule\multicolumn{7}{r}{\textit{Continued on next page}}\\\endfoot
\bottomrule\endlastfoot
Fashion & 2 & 0 & [-0.25, 0.19] & [-0.0052, 0.0066] & [0.962, 0.970] & 0.00284 \\
Fashion & 10 & 8 & [-0.17, 0.11] & [-0.0060, 0.0068] & [0.974, 0.980] & 0.00361 \\
Fashion & 20 & 10 & [-0.25, 0.09] & [-0.0093, 0.0089] & [0.977, 0.984] & 0.00000 \\
KMNIST & 2 & 0 & [-0.19, 0.33] & [-0.0265, 0.0126] & [0.949, 0.960] & 0.00000 \\
KMNIST & 10 & 33 & [-0.14, 0.45] & [-0.0314, 0.0047] & [0.965, 0.969] & 0.00000 \\
KMNIST & 20 & 19 & [-0.38, 0.04] & [-0.0130, 0.0316] & [0.967, 0.970] & 0.00000 \\
MNIST & 2 & 0 & [-0.08, 0.10] & [-0.0059, 0.0031] & [0.962, 0.970] & 0.00000 \\
MNIST & 10 & 45 & [-0.15, 0.09] & [-0.0021, 0.0080] & [0.975, 0.979] & 0.00000 \\
MNIST & 20 & 0 & [-0.03, 0.13] & [-0.0066, 0.0021] & [0.983, 0.987] & 0.00000 \\
\end{longtable}\endgroup

\subsection{Pilot controls, momentum, and batch-order variation}

All 300 pilot perturbations show at least one brief sampled contraction, but all end with separation ratios above one. At strength 20\%, mean accuracy changes from 97.218\% before intervention to 96.607\% immediately afterward and recovers to 97.612\%, compared with 97.567\% for paired baselines. This functional recovery is distinct from meeting the stricter independent-reference thresholds.

\begingroup\footnotesize\setlength{\tabcolsep}{4pt}
\renewcommand{\arraystretch}{1.12}
\begin{longtable}{@{}lrrrrrr@{}}
\caption{Pilot endpoints, five directions per source and ten sources per cell. Distances and functional values are source means; ratio intervals average directions before source bootstrap. Local membership combines proximity to the paired baseline and the reference performance criteria.}\label{tab:s-pilot}\\
\toprule
Momentum & $\epsilon$ (\%) & $r$ [95\% CI] & Accuracy (\%) & Loss & CKA & Local (\%) \\
\midrule\endfirsthead
\multicolumn{7}{l}{\textit{Table \thetable\ continued}}\\
\toprule
Momentum & $\epsilon$ (\%) & $r$ [95\% CI] & Accuracy (\%) & Loss & CKA & Local (\%) \\
\midrule\endhead
\midrule\multicolumn{7}{r}{\textit{Continued on next page}}\\\endfoot
\bottomrule\endlastfoot
preserve & 1 & 16.134 [14.964, 17.294] & 97.599 & 0.12946 & 0.985 & 0 \\
reset & 1 & 15.646 [14.196, 16.916] & 97.610 & 0.12839 & 0.986 & 6 \\
preserve & 5 & 3.344 [3.121, 3.550] & 97.578 & 0.13079 & 0.985 & 0 \\
reset & 5 & 3.305 [3.056, 3.527] & 97.625 & 0.12880 & 0.986 & 0 \\
preserve & 20 & 1.165 [1.128, 1.202] & 97.612 & 0.13113 & 0.984 & 0 \\
reset & 20 & 1.161 [1.120, 1.196] & 97.612 & 0.12998 & 0.984 & 0 \\
\end{longtable}\endgroup

In the moving reference envelope, 136 of 300 immediately perturbed pilot states are already members, and 61 initially outside runs enter later. The archive also records 45 runs retaining membership at every sampled time and 135 sampled entries following an outside observation; these indicators answer different temporal questions and are not additive categories. For the fixed combined region, no perturbed initial state is inside, and 180 runs enter later. None of the ten unperturbed source networks was in that fixed combined final-horizon region before intervention, so these are not perturbation-induced returns. Thirteen of the twenty unperturbed endpoint continuations satisfy its combined criteria. Local endpoint membership occurs only in three of fifty runs at 1\% perturbation with reset momentum; other pilot cells have none.

\begingroup\footnotesize\setlength{\tabcolsep}{4pt}
\renewcommand{\arraystretch}{1.12}
\begin{longtable}{@{}lrrrrr@{}}
\caption{Sensitivity of pilot endpoint membership to the structural quantile and reference ensemble. Entries are percentages [pointwise 95\% bootstrap interval]; reference halves each contain five networks.}\label{tab:s-pilot-regions}\\
\toprule
Momentum & $\epsilon$ (\%) & q95 & q90 & Half A & Half B \\
\midrule\endfirsthead
\multicolumn{6}{l}{\textit{Table \thetable\ continued}}\\
\toprule
Momentum & $\epsilon$ (\%) & q95 & q90 & Half A & Half B \\
\midrule\endhead
\midrule\multicolumn{6}{r}{\textit{Continued on next page}}\\\endfoot
\bottomrule\endlastfoot
preserve & 1 & 72 [48, 92] & 72 [48, 92] & 68 [42, 90] & 52 [30, 74] \\
reset & 1 & 74 [52, 92] & 74 [52, 92] & 64 [40, 84] & 50 [28, 74] \\
preserve & 5 & 62 [38, 84] & 58 [34, 80] & 56 [34, 78] & 50 [28, 72] \\
reset & 5 & 76 [52, 96] & 72 [46, 94] & 72 [46, 94] & 60 [42, 76] \\
preserve & 20 & 34 [12, 58] & 32 [10, 56] & 34 [10, 60] & 42 [18, 66] \\
reset & 20 & 42 [22, 64] & 36 [16, 60] & 44 [18, 72] & 52 [26, 78] \\
\end{longtable}\endgroup

Changing momentum does not reverse the endpoint-expansion pattern. The paired retained-minus-reset ratio intervals all include zero; the table reports the corresponding controls for distances, performance, and region membership as well. A different future batch order also generates separation without an injected weight perturbation: its mean endpoint distance is 3.318 with mean accuracy 97.56\%. The zero-perturbation ratio is undefined, and only its absolute distance is interpreted.

\begingroup\footnotesize\setlength{\tabcolsep}{4pt}
\renewcommand{\arraystretch}{1.12}
\begin{longtable}{@{}rlrr@{}}
\caption{Paired pilot momentum effects: retained minus reset, after averaging directions within source. Accuracy and membership differences are proportions; ratios and distances have their original units. Intervals are pointwise source bootstraps.}\label{tab:s-momentum}\\
\toprule
$\epsilon$ (\%) & Measure & Difference & 95\% interval \\
\midrule\endfirsthead
\multicolumn{4}{l}{\textit{Table \thetable\ continued}}\\
\toprule
$\epsilon$ (\%) & Measure & Difference & 95\% interval \\
\midrule\endhead
\midrule\multicolumn{4}{r}{\textit{Continued on next page}}\\\endfoot
\bottomrule\endlastfoot
1 & ratio & 0.48823 & [-0.25730, 1.26009] \\
1 & distance & 0.08921 & [-0.04705, 0.22988] \\
1 & accuracy & -0.00011 & [-0.00043, 0.00018] \\
1 & loss & 0.00107 & [-0.00038, 0.00231] \\
1 & region member & -0.02000 & [-0.12000, 0.06000] \\
1 & local member & -0.06000 & [-0.18000, 0.00000] \\
5 & ratio & 0.03838 & [-0.06828, 0.13378] \\
5 & distance & 0.03493 & [-0.06238, 0.12222] \\
5 & accuracy & -0.00047 & [-0.00097, -0.00004] \\
5 & loss & 0.00198 & [-0.00039, 0.00457] \\
5 & region member & -0.14000 & [-0.36000, 0.08000] \\
5 & local member & 0.00000 & [0.00000, 0.00000] \\
20 & ratio & 0.00360 & [-0.01672, 0.02257] \\
20 & distance & 0.01332 & [-0.06072, 0.08263] \\
20 & accuracy & 0.00000 & [-0.00043, 0.00036] \\
20 & loss & 0.00115 & [-0.00065, 0.00290] \\
20 & region member & -0.08000 & [-0.20000, 0.02000] \\
20 & local member & 0.00000 & [0.00000, 0.00000] \\
\end{longtable}\endgroup
\begingroup\footnotesize\setlength{\tabcolsep}{4pt}
\renewcommand{\arraystretch}{1.12}
\begin{longtable}{@{}lrrrr@{}}
\caption{Pilot controls changing the future batch sequence. With no injected perturbation, the normalized ratio is undefined because its denominator is zero; the absolute distance measures schedule-induced separation.}\label{tab:s-batch-noise}\\
\toprule
$\epsilon$ (\%) & Runs & Distance [95\% CI] & Accuracy [95\% CI], \% & Loss [95\% CI] \\
\midrule\endfirsthead
\multicolumn{5}{l}{\textit{Table \thetable\ continued}}\\
\toprule
$\epsilon$ (\%) & Runs & Distance [95\% CI] & Accuracy [95\% CI], \% & Loss [95\% CI] \\
\midrule\endhead
\midrule\multicolumn{5}{r}{\textit{Continued on next page}}\\\endfoot
\bottomrule\endlastfoot
0 & 10 & 3.318 [3.132, 3.506] & 97.56 [97.44, 97.70] & 0.1336 [0.1269, 0.1397] \\
20 & 50 & 4.457 [4.328, 4.591] & 97.58 [97.51, 97.66] & 0.1320 [0.1278, 0.1360] \\
\end{longtable}\endgroup

\subsection{Alternative assignments and sign dissimilarity}

Git Re-Basin minimizes the joint canonical Euclidean assignment objective exactly for this architecture. PIF/Hungarian uses compressed neuron features, and Alpha matching optimizes a different correspondence criterion. Each alternative ratio is measured using the canonical joint Frobenius norm after its own assignment. These are sensitivity analyses of the same pairs rather than independent replications of one metric.

\begingroup\footnotesize\setlength{\tabcolsep}{4pt}
\renewcommand{\arraystretch}{1.12}
\begin{longtable}{@{}llrrrr@{}}
\caption{Matching sensitivity across all pilot and expanded cells. Counts classify pointwise 95\% ratio intervals relative to one. These are dependent measurement variants of the same runs, not independent replications.}\label{tab:s-matching}\\
\toprule
Stage & Assignment & Cells & Below 1 & Above 1 & Includes 1 \\
\midrule\endfirsthead
\multicolumn{6}{l}{\textit{Table \thetable\ continued}}\\
\toprule
Stage & Assignment & Cells & Below 1 & Above 1 & Includes 1 \\
\midrule\endhead
\midrule\multicolumn{6}{r}{\textit{Continued on next page}}\\\endfoot
\bottomrule\endlastfoot
Expanded & Git Re-Basin & 216 & 0 & 216 & 0 \\
Expanded & PIF/Hungarian & 216 & 0 & 216 & 0 \\
Expanded & Alpha matching & 216 & 0 & 215 & 1 \\
Pilot & Git Re-Basin & 6 & 0 & 6 & 0 \\
Pilot & PIF/Hungarian & 6 & 0 & 6 & 0 \\
Pilot & Alpha matching & 6 & 0 & 6 & 0 \\
\end{longtable}\endgroup

The sole Alpha-assignment interval including one occurs in Fashion-MNIST at E20, with both matrices perturbed by 20\% and momentum reset: mean 1.095, interval [0.959, 1.198]. No alternative-matching interval lies wholly below one. The separately measured change in weighted sign dissimilarity, $\Delta\alpha$, has intervals below zero in 21 expanded cells, above zero in 184, and including zero in 11. In the pilot the corresponding counts are zero, four, and two. Reduced weighted sign disagreement therefore occurs in some conditions even when Euclidean separation expands; it is not a structural-return criterion. In the repository directory \texttt{06\_Attractor\_Perturbation/}, \href{https://github.com/krausspk/Learning-Trajectory/blob/37b24bc504ffb39c491f28db5c6f18ef820422c2/06_Attractor_Perturbation/alignment_robustness_summary.csv}{\texttt{alignment\_robustness\_summary.csv}} contains the 222 cellwise matching summaries (6 pilot and 216 expanded conditions). The corresponding \href{https://github.com/krausspk/Learning-Trajectory/blob/37b24bc504ffb39c491f28db5c6f18ef820422c2/06_Attractor_Perturbation/alignment_robustness.csv}{\texttt{alignment\_robustness.csv}} contains one row per perturbed pair (300 pilot and 2,160 expanded pairs).

\section{History decoding, layer probes, and symmetry checks (07)}
\label{supp:history}

Both histories of each seed remain in the same train, validation, or test split. The retained-momentum cohort is paired to the reset cohort through the same initialization design; its identical endpoint score is a protocol robustness result. The full probe table includes all horizons, layers, logits, diagnostics, residualized weights, momentum continuation, and the shuffled-label control.

\begingroup\footnotesize\setlength{\tabcolsep}{4pt}
\renewcommand{\arraystretch}{1.12}
\begin{longtable}{@{}lrlrrrr@{}}
\caption{All history probes. Accuracy is balanced accuracy in percent; chance is 50\%. Intervals resample the 40 held-out seed pairs. The primary endpoint is reset/both/C10; Holm probabilities concern the specified secondary family. Dashes indicate that a probability was not defined for that row.}\label{tab:s-history}\\
\toprule
Momentum & C epochs & Features & Accuracy & 95\% interval & $p$ & Holm $p$ \\
\midrule\endfirsthead
\multicolumn{7}{l}{\textit{Table \thetable\ continued}}\\
\toprule
Momentum & C epochs & Features & Accuracy & 95\% interval & $p$ & Holm $p$ \\
\midrule\endhead
\midrule\multicolumn{7}{r}{\textit{Continued on next page}}\\\endfoot
\bottomrule\endlastfoot
reset & 1 & $W_0$ & 100.00 & [100.00, 100.00] & 0.001 & 0.019 \\
reset & 1 & $W_1$ & 100.00 & [100.00, 100.00] & 0.001 & 0.019 \\
reset & 1 & Both weights & 100.00 & [100.00, 100.00] & 0.001 & 0.019 \\
reset & 1 & Diagnostics & 85.00 & [77.50, 92.50] & 0.001 & 0.019 \\
reset & 1 & Logits & 96.25 & [91.25, 100.00] & 0.001 & 0.019 \\
reset & 1 & Residual weights & 83.75 & [75.00, 91.25] & 0.001 & 0.019 \\
reset & 5 & $W_0$ & 100.00 & [100.00, 100.00] & 0.001 & 0.019 \\
reset & 5 & $W_1$ & 100.00 & [100.00, 100.00] & 0.001 & 0.019 \\
reset & 5 & Both weights & 100.00 & [100.00, 100.00] & 0.001 & 0.019 \\
reset & 5 & Diagnostics & 87.50 & [80.00, 93.75] & 0.001 & 0.019 \\
reset & 5 & Logits & 90.00 & [83.75, 96.25] & 0.001 & 0.019 \\
reset & 5 & Residual weights & 75.00 & [66.25, 83.75] & 0.001 & 0.019 \\
reset & 10 & $W_0$ & 100.00 & [100.00, 100.00] & 0.001 & 0.019 \\
reset & 10 & $W_1$ & 100.00 & [100.00, 100.00] & 0.001 & 0.019 \\
reset & 10 & Both weights & 100.00 & [100.00, 100.00] & 0.001 & -- \\
reset & 10 & Diagnostics & 83.75 & [76.25, 90.00] & 0.001 & 0.019 \\
reset & 10 & Logits & 85.00 & [77.50, 92.50] & 0.001 & 0.019 \\
reset & 10 & Residual weights & 81.25 & [72.50, 88.75] & 0.001 & 0.019 \\
reset & 10 & Both weights (shuffled) & 48.75 & [40.00, 58.75] & 0.617 & 0.617 \\
continued & 10 & Both weights & 100.00 & [100.00, 100.00] & 0.001 & 0.019 \\
\end{longtable}\endgroup

Both $W_0$ and $W_1$ probes achieve 100\% balanced accuracy at C1, C5, and C10. Their features follow joint gauge balancing, normalization, and neuron assignment, so this does not demonstrate independent decoding from each isolated raw matrix. It is also a different target from the output-layer initialization retrieval in experiment 03: a layer contribution can classify earlier task order even when it poorly retrieves a specific known initialization under a different comparison.

Symmetry validation reorders hidden units and applies positive ReLU rescalings before realignment. Across 1,200 checked configurations per momentum protocol, the maximum aligned-feature change after permutation is zero. Maximum rescaling errors are $3.55\times10^{-15}$ for continued momentum and $3.11\times10^{-15}$ for reset momentum. Thus, the aligned feature representation passes the explicit numerical symmetry checks.

\begingroup\footnotesize\setlength{\tabcolsep}{4pt}
\renewcommand{\arraystretch}{1.12}
\begin{longtable}{@{}llrr@{}}
\caption{Final-task differences on held-out history seeds: B--A--C minus A--B--C. Accuracy differences and intervals are in percentage points; loss remains cross-entropy. Forty seed pairs per condition.}\label{tab:s-history-function}\\
\toprule
Momentum & Measure & Difference & 95\% interval \\
\midrule\endfirsthead
\multicolumn{4}{l}{\textit{Table \thetable\ continued}}\\
\toprule
Momentum & Measure & Difference & 95\% interval \\
\midrule\endhead
\midrule\multicolumn{4}{r}{\textit{Continued on next page}}\\\endfoot
\bottomrule\endlastfoot
reset & accuracy & -0.8065 & [-1.1788, -0.4812] \\
reset & loss & 0.0246 & [0.0124, 0.0374] \\
continued & accuracy & -0.8170 & [-1.1280, -0.5179] \\
continued & loss & 0.0275 & [0.0171, 0.0380] \\
\end{longtable}\endgroup

Logits, performance diagnostics, and residualized weights carry history information as reported in the main text. Residualization removes only train-fitted linear relationships with the six measured diagnostics. Raw-logit decoding is not the same as decoding probabilities or predicted classes: a common shift of logits can leave softmax probabilities unchanged. The performance difference independently establishes a functional difference in this cohort; no equivalence of the two final predictors is assumed. Task order also determines which task immediately precedes C, so those two historical properties are not independently manipulated here.

\section{Trajectory and event controls (08)}
\label{supp:trajectory}

\subsection{Reconstruction targets and decoder robustness}

The two new cohorts contain 200 seed pairs each, split 120/40/40. The one-pair and four-pair curricula end with five C epochs. A separate exploratory analysis of the original 30 networks served pipeline development and is not another independent validation cohort. Both earlier target coordinates and endpoint feature preprocessing are learned using the specified training split; earlier test states enter only as targets. The endpoint-derived permutation is fixed across a network's trajectory, and the known endpoint is excluded from reconstruction scores.

Absolute initial-state prediction remains weak when increasing PCA dimension, changing layer features, or using the nonlinear probe. The MLP operates on a training-endpoint PCA embedding with up to 32 coordinates and has 32 hidden units; validation chooses the saved step within 300 optimization steps. It does not improve initial-state reconstruction. The slightly positive $R^2$ for the four-pair output-layer, eight-coordinate condition is approximately 0.003, not substantial individualized recovery.

\begingroup\footnotesize\setlength{\tabcolsep}{4pt}
\renewcommand{\arraystretch}{1.12}
\begin{longtable}{@{}llrlrrr@{}}
\caption{Initial-state reconstruction across all planned layer/PCA variants and the nonlinear probe. Targets are absolute earlier coordinates; the endpoint is never a target checkpoint. MSE and $R^2$ describe the same held-out seed blocks. No row provides accurate individualized start reconstruction.}\label{tab:s-start-robustness}\\
\toprule
Cohort & Layer & PCA dim. & Probe & $R^2$ & MSE & MSE 95\% interval \\
\midrule\endfirsthead
\multicolumn{7}{l}{\textit{Table \thetable\ continued}}\\
\toprule
Cohort & Layer & PCA dim. & Probe & $R^2$ & MSE & MSE 95\% interval \\
\midrule\endhead
\midrule\multicolumn{7}{r}{\textit{Continued on next page}}\\\endfoot
\bottomrule\endlastfoot
One-pair & Both & 8 & RIDGE & -0.199 & 0.06937 & [0.06189, 0.07681] \\
One-pair & Both & 16 & RIDGE & -0.079 & 0.06622 & [0.06170, 0.07029] \\
One-pair & Both & 32 & RIDGE & -0.040 & 0.06229 & [0.05921, 0.06521] \\
One-pair & Both & 8 & MLP & -1.869 & 0.16599 & [0.14798, 0.18650] \\
One-pair & $W_0$ & 8 & RIDGE & -0.200 & 0.06940 & [0.06191, 0.07684] \\
One-pair & $W_0$ & 16 & RIDGE & -0.080 & 0.06627 & [0.06174, 0.07033] \\
One-pair & $W_0$ & 32 & RIDGE & -0.040 & 0.06233 & [0.05924, 0.06523] \\
One-pair & $W_1$ & 8 & RIDGE & -0.205 & 0.06971 & [0.06181, 0.07763] \\
One-pair & $W_1$ & 16 & RIDGE & -0.090 & 0.06690 & [0.06155, 0.07186] \\
One-pair & $W_1$ & 32 & RIDGE & -0.070 & 0.06413 & [0.06071, 0.06734] \\
Four-pair & Both & 8 & RIDGE & -0.009 & 0.02958 & [0.02639, 0.03283] \\
Four-pair & Both & 16 & RIDGE & -0.040 & 0.03218 & [0.02992, 0.03440] \\
Four-pair & Both & 32 & RIDGE & -0.035 & 0.03196 & [0.03028, 0.03380] \\
Four-pair & Both & 8 & MLP & -0.577 & 0.04624 & [0.04094, 0.05171] \\
Four-pair & $W_0$ & 8 & RIDGE & -0.009 & 0.02960 & [0.02642, 0.03286] \\
Four-pair & $W_0$ & 16 & RIDGE & -0.041 & 0.03220 & [0.02994, 0.03441] \\
Four-pair & $W_0$ & 32 & RIDGE & -0.035 & 0.03197 & [0.03029, 0.03382] \\
Four-pair & $W_1$ & 8 & RIDGE & 0.003 & 0.02924 & [0.02574, 0.03266] \\
Four-pair & $W_1$ & 16 & RIDGE & -0.040 & 0.03216 & [0.02963, 0.03449] \\
Four-pair & $W_1$ & 32 & RIDGE & -0.057 & 0.03266 & [0.03096, 0.03443] \\
\end{longtable}\endgroup
\begingroup\footnotesize\setlength{\tabcolsep}{4pt}
\renewcommand{\arraystretch}{1.12}
\begin{longtable}{@{}lrr@{}}
\caption{Fraction of scaled training-trajectory variance captured by each target PCA basis. The basis is learned from training trajectories only.}\label{tab:s-pca}\\
\toprule
Cohort & Coordinates & Explained variance (\%) \\
\midrule\endfirsthead
\multicolumn{3}{l}{\textit{Table \thetable\ continued}}\\
\toprule
Cohort & Coordinates & Explained variance (\%) \\
\midrule\endhead
\midrule\multicolumn{3}{r}{\textit{Continued on next page}}\\\endfoot
\bottomrule\endlastfoot
One-pair & 8 & 27.57 \\
One-pair & 16 & 32.07 \\
One-pair & 32 & 38.14 \\
Four-pair & 8 & 28.33 \\
Four-pair & 16 & 33.09 \\
Four-pair & 32 & 39.85 \\
\end{longtable}\endgroup

The primary displacement $R^2$ values of approximately 0.986 and 0.953 are exceeded by the mean-early-minus-endpoint control (approximately 0.9998 in both cohorts). That control has exactly the absolute-coordinate mean-baseline MSE because it subtracts the same known endpoint from target and prediction. The high displacement score therefore cannot be interpreted as accurate start reconstruction. Likewise, the significant target-correspondence permutation result in the four-pair primary analysis tests matching of individual target trajectories; it does not establish superiority to the training-mean baseline. Full checkpoint curves and all baselines are supplied in the numerical archive.

\begingroup\footnotesize\setlength{\tabcolsep}{4pt}
\renewcommand{\arraystretch}{1.12}
\begin{longtable}{@{}llrrr@{}}
\caption{Random-coordinate negative controls. Independent random targets are fitted and evaluated with the regression pipeline; the mean predictor uses training targets only. Individual random-control realizations need not equal zero $R^2$.}\label{tab:s-random-target}\\
\toprule
Cohort & Probe & $R^2$ & 95\% interval & MSE \\
\midrule\endfirsthead
\multicolumn{5}{l}{\textit{Table \thetable\ continued}}\\
\toprule
Cohort & Probe & $R^2$ & 95\% interval & MSE \\
\midrule\endhead
\midrule\multicolumn{5}{r}{\textit{Continued on next page}}\\\endfoot
\bottomrule\endlastfoot
One-pair & Ridge & -0.0272 & [-0.0762, -0.0166] & 0.30868 \\
One-pair & Mean & -0.0155 & [-0.0595, -0.0097] & 0.30516 \\
Four-pair & Ridge & -0.0082 & [-0.0484, -0.0018] & 0.09759 \\
Four-pair & Mean & -0.0143 & [-0.0543, -0.0099] & 0.09819 \\
\end{longtable}\endgroup

\subsection{All event probes and functional controls}

The full table includes six linear feature conditions per event position, the nonlinear probe, and shuffled-label controls. Feature-by-position Holm correction is applied within each cohort's prespecified linear family. The earliest position is the primary event endpoint. In the four-pair cohort, only the first two joint-weight positions survive that correction. The MLP retains perfect one-pair classification but yields 81.25\%, 50\%, 53.75\%, and 58.75\% from oldest to newest in the four-pair cohort. Shuffled-label accuracies remain around chance.

\begingroup\footnotesize\setlength{\tabcolsep}{4pt}
\renewcommand{\arraystretch}{1.12}
\begin{longtable}{@{}lrllrrr@{}}
\caption{All event probes, including layer features, performance/norm controls, nonlinear probes, and shuffled-label controls. Positions are numbered oldest to newest. Accuracy intervals are seed-block bootstraps. Holm correction applies to the prespecified linear feature-by-position family in each cohort; no adjusted probabilities were computed for MLP or shuffled-label rows.}\label{tab:s-events}\\
\toprule
Cohort & Position & Features & Probe & Accuracy (\%) & 95\% interval & Holm $p$ \\
\midrule\endfirsthead
\multicolumn{7}{l}{\textit{Table \thetable\ continued}}\\
\toprule
Cohort & Position & Features & Probe & Accuracy (\%) & 95\% interval & Holm $p$ \\
\midrule\endhead
\midrule\multicolumn{7}{r}{\textit{Continued on next page}}\\\endfoot
\bottomrule\endlastfoot
One-pair & 1 & Both & Ridge & 100.00 & [100.00, 100.00] & 0.006 \\
One-pair & 1 & Both & Shuffled & 55.00 & [38.75, 68.78] & -- \\
One-pair & 1 & Both & MLP & 100.00 & [100.00, 100.00] & -- \\
One-pair & 1 & W0 & Ridge & 100.00 & [100.00, 100.00] & 0.006 \\
One-pair & 1 & W1 & Ridge & 100.00 & [100.00, 100.00] & 0.006 \\
One-pair & 1 & performance & Ridge & 70.00 & [61.25, 78.75] & 0.006 \\
One-pair & 1 & norms & Ridge & 75.00 & [67.50, 83.75] & 0.006 \\
One-pair & 1 & diagnostics & Ridge & 87.50 & [81.22, 93.75] & 0.006 \\
Four-pair & 1 & Both & Ridge & 93.75 & [88.75, 98.75] & 0.024 \\
Four-pair & 2 & Both & Ridge & 68.75 & [60.00, 77.50] & 0.040 \\
Four-pair & 3 & Both & Ridge & 63.75 & [53.75, 73.75] & 0.216 \\
Four-pair & 4 & Both & Ridge & 60.00 & [48.75, 70.03] & 0.600 \\
Four-pair & 1 & Both & Shuffled & 47.50 & [36.25, 57.50] & -- \\
Four-pair & 2 & Both & Shuffled & 48.75 & [38.75, 58.75] & -- \\
Four-pair & 3 & Both & Shuffled & 52.50 & [43.75, 61.25] & -- \\
Four-pair & 4 & Both & Shuffled & 52.50 & [42.50, 62.50] & -- \\
Four-pair & 1 & Both & MLP & 81.25 & [72.50, 90.00] & -- \\
Four-pair & 2 & Both & MLP & 50.00 & [41.25, 60.00] & -- \\
Four-pair & 3 & Both & MLP & 53.75 & [42.50, 66.25] & -- \\
Four-pair & 4 & Both & MLP & 58.75 & [48.75, 67.50] & -- \\
Four-pair & 1 & W0 & Ridge & 93.75 & [88.75, 98.75] & 0.024 \\
Four-pair & 2 & W0 & Ridge & 70.00 & [61.25, 78.75] & 0.024 \\
Four-pair & 3 & W0 & Ridge & 63.75 & [53.75, 73.75] & 0.095 \\
Four-pair & 4 & W0 & Ridge & 58.75 & [47.50, 68.75] & 1.000 \\
Four-pair & 1 & W1 & Ridge & 95.00 & [90.00, 98.75] & 0.024 \\
Four-pair & 2 & W1 & Ridge & 60.00 & [50.00, 70.00] & 0.742 \\
Four-pair & 3 & W1 & Ridge & 58.75 & [47.50, 68.75] & 1.000 \\
Four-pair & 4 & W1 & Ridge & 52.50 & [42.50, 62.53] & 1.000 \\
Four-pair & 1 & performance & Ridge & 51.25 & [40.00, 61.25] & 1.000 \\
Four-pair & 2 & performance & Ridge & 51.25 & [42.50, 60.00] & 1.000 \\
Four-pair & 3 & performance & Ridge & 60.00 & [50.00, 70.03] & 0.432 \\
Four-pair & 4 & performance & Ridge & 43.75 & [34.97, 53.75] & 1.000 \\
Four-pair & 1 & norms & Ridge & 46.25 & [36.22, 57.50] & 1.000 \\
Four-pair & 2 & norms & Ridge & 46.25 & [36.25, 56.28] & 1.000 \\
Four-pair & 3 & norms & Ridge & 51.25 & [40.00, 62.50] & 1.000 \\
Four-pair & 4 & norms & Ridge & 55.00 & [45.00, 66.25] & 1.000 \\
Four-pair & 1 & diagnostics & Ridge & 48.75 & [36.25, 60.00] & 1.000 \\
Four-pair & 2 & diagnostics & Ridge & 48.75 & [37.50, 60.00] & 1.000 \\
Four-pair & 3 & diagnostics & Ridge & 55.00 & [43.75, 65.03] & 1.000 \\
Four-pair & 4 & diagnostics & Ridge & 62.50 & [51.25, 73.75] & 0.306 \\
\end{longtable}\endgroup

One-pair final C accuracy averages 92.848\%, with range 85.90--95.29\%, whereas four-pair accuracy averages 29.888\%, with range 9.24--86.25\%. The full-cohort one-pair history difference is $-1.086$ percentage points (bit 1 minus bit 0); the held-out difference is $-0.949$ points. Performance and norm features consequently carry some of the one-pair event information. In the four-pair cohort, no run reaches the descriptive 90\% final-accuracy threshold. Mean hidden-unit inactivity is 61.16 of 64 at the endpoint and 60.83 immediately before C on the fixed 1,024-image probe. All runs remain in the analyses. These diagnostics identify a failed final-learning regime but do not establish the causal mechanism of that failure.

\begingroup\footnotesize\setlength{\tabcolsep}{4pt}
\renewcommand{\arraystretch}{1.12}
\begin{longtable}{@{}llrlrr@{}}
\caption{Paired final-performance differences for event cohorts. All = 200 seed pairs; test = 40 held-out pairs. Bit 1 minus bit 0 refers to complementary order labels. Accuracy differences are percentage points; loss is cross-entropy. These are descriptive intervals, not equivalence tests or isolated causal effects of one bit.}\label{tab:s-event-function}\\
\toprule
Cohort & Split & Position & Measure & Difference & 95\% interval \\
\midrule\endfirsthead
\multicolumn{6}{l}{\textit{Table \thetable\ continued}}\\
\toprule
Cohort & Split & Position & Measure & Difference & 95\% interval \\
\midrule\endhead
\midrule\multicolumn{6}{r}{\textit{Continued on next page}}\\\endfoot
\bottomrule\endlastfoot
One-pair & all & 1 & accuracy & -1.086 & [-1.309, -0.877] \\
One-pair & all & 1 & loss & 0.036 & [0.029, 0.043] \\
One-pair & test & 1 & accuracy & -0.949 & [-1.341, -0.564] \\
One-pair & test & 1 & loss & 0.034 & [0.020, 0.049] \\
Four-pair & all & 1 & accuracy & -0.459 & [-3.924, 3.010] \\
Four-pair & all & 1 & loss & 0.015 & [-0.064, 0.092] \\
Four-pair & all & 2 & accuracy & -1.307 & [-4.659, 2.131] \\
Four-pair & all & 2 & loss & 0.044 & [-0.032, 0.125] \\
Four-pair & all & 3 & accuracy & -2.722 & [-6.243, 0.842] \\
Four-pair & all & 3 & loss & 0.061 & [-0.014, 0.139] \\
Four-pair & all & 4 & accuracy & -1.518 & [-5.050, 1.815] \\
Four-pair & all & 4 & loss & 0.027 & [-0.056, 0.104] \\
Four-pair & test & 1 & accuracy & 2.275 & [-5.584, 10.021] \\
Four-pair & test & 1 & loss & -0.033 & [-0.221, 0.142] \\
Four-pair & test & 2 & accuracy & -3.651 & [-12.086, 4.172] \\
Four-pair & test & 2 & loss & 0.077 & [-0.111, 0.257] \\
Four-pair & test & 3 & accuracy & -7.238 & [-14.763, -0.234] \\
Four-pair & test & 3 & loss & 0.160 & [0.003, 0.324] \\
Four-pair & test & 4 & accuracy & -0.315 & [-8.419, 7.740] \\
Four-pair & test & 4 & loss & 0.015 & [-0.161, 0.191] \\
\end{longtable}\endgroup

Perfect per-network accuracy yields a degenerate empirical bootstrap interval in the one-pair cohort. A separate exact binomial calculation treats an entire held-out seed pair as a success only if both histories are correct. All 40 one-pair seeds succeed, giving a 95\% interval of 91.19--100\% for this stricter pair-success probability conditional on the fitted probe. In the four-pair cohort, only one seed pair has all four positions correctly decoded in both arms. This is not the per-network four-bit accuracy; it is a stricter, explicitly defined endpoint that illustrates the difference between reading selected events and recovering the complete tested history.

\begingroup\footnotesize\setlength{\tabcolsep}{4pt}
\renewcommand{\arraystretch}{1.12}
\begin{longtable}{@{}llrrr@{}}
\caption{Whole-seed-pair success, a stricter outcome than per-network event accuracy. A success requires correct prediction for both histories of a seed; the all-bits row additionally requires all four positions. Exact binomial intervals condition on the fitted probe.}\label{tab:s-pair-success}\\
\toprule
Cohort & Position & Successful pairs & Rate (\%) & Exact 95\% interval \\
\midrule\endfirsthead
\multicolumn{5}{l}{\textit{Table \thetable\ continued}}\\
\toprule
Cohort & Position & Successful pairs & Rate (\%) & Exact 95\% interval \\
\midrule\endhead
\midrule\multicolumn{5}{r}{\textit{Continued on next page}}\\\endfoot
\bottomrule\endlastfoot
One-pair & 1 & 40/40 & 100.0 & [91.19, 100.00] \\
Four-pair & 1 & 35/40 & 87.5 & [73.20, 95.81] \\
Four-pair & 2 & 17/40 & 42.5 & [27.04, 59.11] \\
Four-pair & 3 & 16/40 & 40.0 & [24.86, 56.67] \\
Four-pair & 4 & 14/40 & 35.0 & [20.63, 51.68] \\
Four-pair & All four & 1/40 & 2.5 & [0.06, 13.16] \\
\end{longtable}\endgroup

\subsection{Reproducibility and coverage}

The full analysis archive retains numerical results, generation scripts, source hashes, and original reports, including inferential quantities and checkpoint controls omitted from the printed tables. This supplement covers all eight experiments and the complete perturbation extension. The submission project contains self-contained main and supplementary LaTeX documents, the bibliography, and figure assets; it compiles independently of the separate analysis archive.

\end{document}